\documentclass[]{semocoarxiv}

\usepackage{algorithm}
\usepackage{algorithmic}
\usepackage{amsmath}
\usepackage{amssymb}
\usepackage{amsfonts}
\usepackage{adjustbox}
\usepackage{flafter}

\let\cite\citep

\title{SeMoCo: A Semantic-First Motion Codec for Motion Language Modeling}

\author[1,2,*]{Tianlv Huang}
\author[1,2,*]{Hetian Guo}
\author[3]{Ziyi Cai}
\author[2]{Song Wang}
\author[2]{Yanping Zhang}
\author[1,\dagger]{Zipei Fan}
\author[1]{Xuan Song}
\author[2,\ddagger]{Guangming Wu}
\author[2,\ddagger]{Xin Zheng}

\affiliation[1]{Jilin University}
\affiliation[2]{Frontier Robotics}
\affiliation[3]{Harbin Institute of Technology, Shenzhen}
\contribution[*]{Equal contribution}
\contribution[\dagger]{Corresponding author}
\contribution[\ddagger]{Project leads}

\date{\today}
\metadata[Tokenizer code]{\url{https://github.com/OMEGA-i/SeMoCo-Tokenizer}}
\metadata[Generator code]{\url{https://github.com/OMEGA-i/SeMoCo-Generator}}

\def\omvfigwidth{0.42}
\def\mtfigwidth{0.68}

\begin{document}

\begin{abstract}
Discrete motion representations have substantially advanced autoregressive
text-to-motion generation. However, most motion tokenizers are optimized for
reconstruction and do not explicitly allocate capacity according to semantic
role. Action-level meaning and fine-grained kinematic detail must therefore be
encoded through the same reconstruction-driven hierarchy. We
introduce SeMoCo, a semantic-first motion codec, together with a dual-axis
motion generator for language-conditioned motion generation. Each motion token
contains one semantic token and a residual sequence of kinematic tokens. The
generator models semantic progression across time and autoregressively refines
the residual entries. We also construct $\Omega$-MotionVerse, a large-scale,
multi-source human-motion dataset unified under the SOMA representation.
Across the reported comparisons, SeMoCo achieves the best reconstruction
accuracy among the compared codecs, while strong text-to-motion results
demonstrate the effectiveness of its motion tokens for downstream generation.
\end{abstract}

\maketitle
\section{Introduction}
Text-to-motion (T2M) generation, which synthesizes realistic and temporally
coherent human motion from natural-language descriptions, has advanced through
both continuous and discrete generative models. Continuous approaches generate
poses or motion latents with diffusion and flow-based models
~\cite{tevet2022humanmotiondiffusionmodel,
chen2023executingcommandsmotiondiffusion,wen2025hymotion,rempe2026kimodo},
whereas discrete approaches encode motion into learned tokens and model the
resulting sequences with autoregressive or masked generators
~\cite{zhang2023t2mgpt,jiang2023motiongpt,guo2024momask,fu2026mogo}. Language
models have shown that discrete tokens can support scalable sequence
generation, and neural audio models extend this recipe to continuous waveforms
with codec tokens~\cite{borsos2023audiolmlanguagemodelingapproach,
wang2023neuralcodeclanguagemodels}.

Directly transferring this paradigm to motion is nontrivial because text
specifies action intent and coarse temporal structure, whereas a full-body
sequence must realize that intent through coordinated trajectories, contacts,
articulation, and smooth dynamics. Most motion tokenizers nevertheless optimize
VQ/RVQ representations primarily for reconstruction, organizing later
codebooks by successive reconstruction residuals rather than explicit motion
semantics~\cite{oord2018neuraldiscreterepresentationlearning,
zeghidour2021soundstreamendtoendneuralaudio}. Speech tokenizers have explored a
semantic-first organization by distilling semantic information from a
self-supervised speech model into the first RVQ level, while later levels retain
residual paralinguistic detail
~\cite{zhang2024speechtokenizerunifiedspeechtokenizer}. Moshi's Mimi codec
further reports a trade-off between semantic discriminability and reconstruction
in this single-RVQ design and instead uses a semantic VQ in parallel with an
acoustic RVQ~\cite{defossez2024moshispeechtextfoundationmodel}.

We therefore introduce \textbf{SeMoCo}, a semantic-first motion codec that
distills window-level semantic knowledge from a frozen Text-to-Motion Retrieval
(TMR) encoder into a dedicated semantic VQ~\cite{petrovich2023tmr}, while a
parallel RVQ encodes residual kinematic detail. Their quantized outputs jointly
form each motion token and are decoded for reconstruction. Building on this token structure, our motion generator performs hierarchical
semantic-to-kinematic token generation by predicting semantic tokens across
time with a temporal Transformer and autoregressively completing residual
kinematic tokens at each position with a lightweight refinement decoder. This
factorization keeps temporal modeling at the motion-token level and confines
autoregressive refinement to the local kinematic hierarchy. We further construct $\Omega$-MotionVerse, a large-scale, multi-source corpus
comprising roughly 1,000 hours of text-annotated human motion standardized to
the SOMA skeleton convention~\cite{saito2026somaunifyingparametrichuman}.

Across motion reconstruction, text-to-motion generation, and motion prediction,
SeMoCo achieves the best reconstruction accuracy among the compared codecs,
while strong text-to-motion results demonstrate the effectiveness of its motion
tokens for downstream generation.

\noindent\textbf{Our contributions are threefold:} (1) \textbf{SeMoCo}, a
semantic-first motion codec that gives semantic and kinematic entries separate
supervision and quantization paths within each motion token; (2) a dual-axis motion generator that models semantic progression across time
and autoregressive kinematic refinement within each motion token; and (3)
$\Omega$-MotionVerse, a large-scale, multi-source human-motion dataset unified
under the SOMA representation.

\section{Related Work}

\subsection{Semantic Motion Tokenization}
Most discrete motion pipelines learn this interface from reconstruction. VQ
maps continuous features to categorical latents, whereas RVQ successively
encodes the remaining reconstruction error~\cite{oord2018neuraldiscreterepresentationlearning,
defossez2022highfidelityneuralaudio,zhang2023t2mgpt,guo2024momask}. Under
reconstruction-only training, the resulting hierarchy is ordered by distortion
reduction rather than by an explicit semantic target. Prior work introduces
semantic structure at different points in the pipeline. TMR and MoLingo shape
text--motion embedding or continuous latent spaces, whereas LG-Tok conditions a
discrete tokenizer and detokenizer on language~\cite{petrovich2023tmr,
he2025molingo,yan2026lgtok}. PGR$^2$M takes a different route by placing
predefined, interpretable pose codes before learned residual refinement, while
Latent Motion Reasoning (LMR) learns separate semantic-reasoning and
motion-execution sequences~\cite{jeong2025poseguidedresidualrefinementinterpretable,
qian2025thinkmovelatentmotion}. MoGeFlow instead exploits the geometry of a
frozen codebook during generation, while Beyond MoCap studies how data and
codebook scale affect tokenization~\cite{fang2026mogeflow,yan2026beyondmocap}.
These methods differ in where semantics enters and how it is paired with motion
detail.

The closest architectural precedents to SeMoCo come from speech. Mimi,
introduced with Moshi, places a teacher-distilled semantic VQ alongside an
acoustic RVQ, and Qwen3-TTS follows the same split-codec
principle~\cite{defossez2024moshispeechtextfoundationmodel,
hu2026qwen3ttstechnicalreport}. Inspired by these systems, SeMoCo studies the
motion-specific form of this design: a teacher-aligned semantic code and a
complete kinematic RVQ path share each temporal packet and reconstruction
decoder, encouraging role specialization without assuming strict
disentanglement.

\subsection{Hierarchical Multi-Codebook Generation}
Assigning several codes to each time step introduces dependencies both across
time and within the representational hierarchy. Motion models have largely
organized this hierarchy by residual depth or scale: MoMask and MOGO operate
over residual codebook levels, whereas MoSa, MoScale, and ScaleMoGen proceed
coarse-to-fine across temporal or skeletal--temporal
scales~\cite{guo2024momask,fu2026mogo,liu2025mosa,
zheng2026nextscaleautoregressivemodelstexttomotion,hwang2026scalemogen}. Audio
generation exposes a complementary design space. MusicGen studies alternative
serialization and interleaving patterns, while Moshi and Qwen3-TTS separate
long-range temporal modeling from within-frame code
completion~\cite{copet2024simplecontrollablemusicgeneration,
defossez2024moshispeechtextfoundationmodel,hu2026qwen3ttstechnicalreport}.
SeMoCo is directly inspired by this temporal--depth factorization and applies
it to semantic-first motion packets. Its focus is the coupling between an
established generation order and a separately supervised motion
representation, rather than the factorization itself.

\subsection{Text-to-Motion Generation}
Text-to-motion methods differ not only in their generative objectives, but also
in the representation over which those objectives are learned. Continuous
approaches synthesize pose trajectories or learned latents with diffusion and
flow objectives, ranging from MotionDiffuse, the Human Motion Diffusion Model
(MDM), and Motion Latent Diffusion (MLD) to scaled systems such as HY-Motion and
controllable whole-body models such as Kimodo~\cite{zhang2022motiondiffuse,
tevet2022humanmotiondiffusionmodel,chen2023executingcommandsmotiondiffusion,
wen2025hymotion,rempe2026kimodo,wang2026motionbricks}. Related efforts have also scaled motion data
and autoregressive models to broaden zero-shot generation~\cite{fan2025go}.
Discrete approaches instead learn a motion vocabulary and model its codes:
T2M-GPT and MotionGPT treat motion as a single token stream, while MoMask, MOGO,
and MoScale organize prediction through residual or multi-scale
structures~\cite{zhang2023t2mgpt,jiang2023motiongpt,guo2024momask,fu2026mogo,
zheng2026nextscaleautoregressivemodelstexttomotion}. This progression makes the
tokenizer more than a compression front end: its code organization defines the
information available at each prediction step and the dependencies that the
generator must resolve.

\section{Method}
\label{sec:method}

SeMoCo is a semantic-first motion codec that combines semantic and kinematic
information through motion--language distillation and residual vector
quantization. It represents motion as semantic-to-kinematic packets, each
containing a semantically aligned primary code and residual codes for
fine-grained kinematic details. A dual-axis motion language model then performs
language-conditioned generation by modeling semantic progression across
packets and predicting kinematic residuals within each packet.

\subsection{Motion Representation}
\label{sec:motion-representation}

Let $\mathbf{x}_{1:N}$ denote a whole-body motion sequence sampled at 50 Hz.
We canonicalize the sequence by aligning it with the floor and removing its
initial planar translation and heading. The removed global state is stored in
a clip-level anchor $\mathbf{a}$, while the remaining motion is represented as
transition records
\begin{align}
  \mathcal{U}(\mathbf{x}_{1:N})
  &= \left(\mathbf{a},\mathbf{u}_{1:N-1}\right),
  \nonumber\\
  \mathbf{u}_n
  &= \left[
    \mathbf{r}^{\mathrm{traj}}_n,
    \mathbf{r}^{\mathrm{root}}_n,
    \mathbf{r}^{\mathrm{joint}}_n,
    \mathbf{v}^{\mathrm{sparse}}_n,
    \mathbf{c}^{\mathrm{foot}}_n
  \right] \in \mathbb{R}^{d_u}.
  \label{eq:umr-representation}
\end{align}
The five feature groups describe the root trajectory, root orientation,
parent-local joint rotations, sparse velocities at contact-relevant joints,
and foot-contact states, respectively~\cite{wang2026motionbricks}.

Planar root translation is encoded as frame-to-frame displacements in both
root-yaw-local and canonical world frames, then averaged and integrated
during recovery to preserve continuity and mitigate drift. Root height,
orientation, and local joint rotations remain absolute, requiring no
temporal integration. This controlled redundancy supports accurate and
robust trajectory reconstruction. Given decoded records $\widehat{\mathbf{u}}_{1:N-1}$, the full-body motion is
recovered as
\begin{equation}
  \widehat{\mathbf{x}}_{1:N}
  = \operatorname{FK}\!\left(
      \mathcal{M}\!\left(
        \mathbf{a},\widehat{\mathbf{u}}_{1:N-1}
      \right)
    \right),
  \label{eq:umr-materialization}
\end{equation}
where $\mathcal{M}$ restores the global trajectory and
$\operatorname{FK}$ applies forward kinematics. Reconstruction and motion
prediction retain the anchor of the observed sequence, while
text-conditioned generation starts from a fixed canonical anchor.
\begin{figure*}[t]
  \centering
  \includegraphics[width=\textwidth]{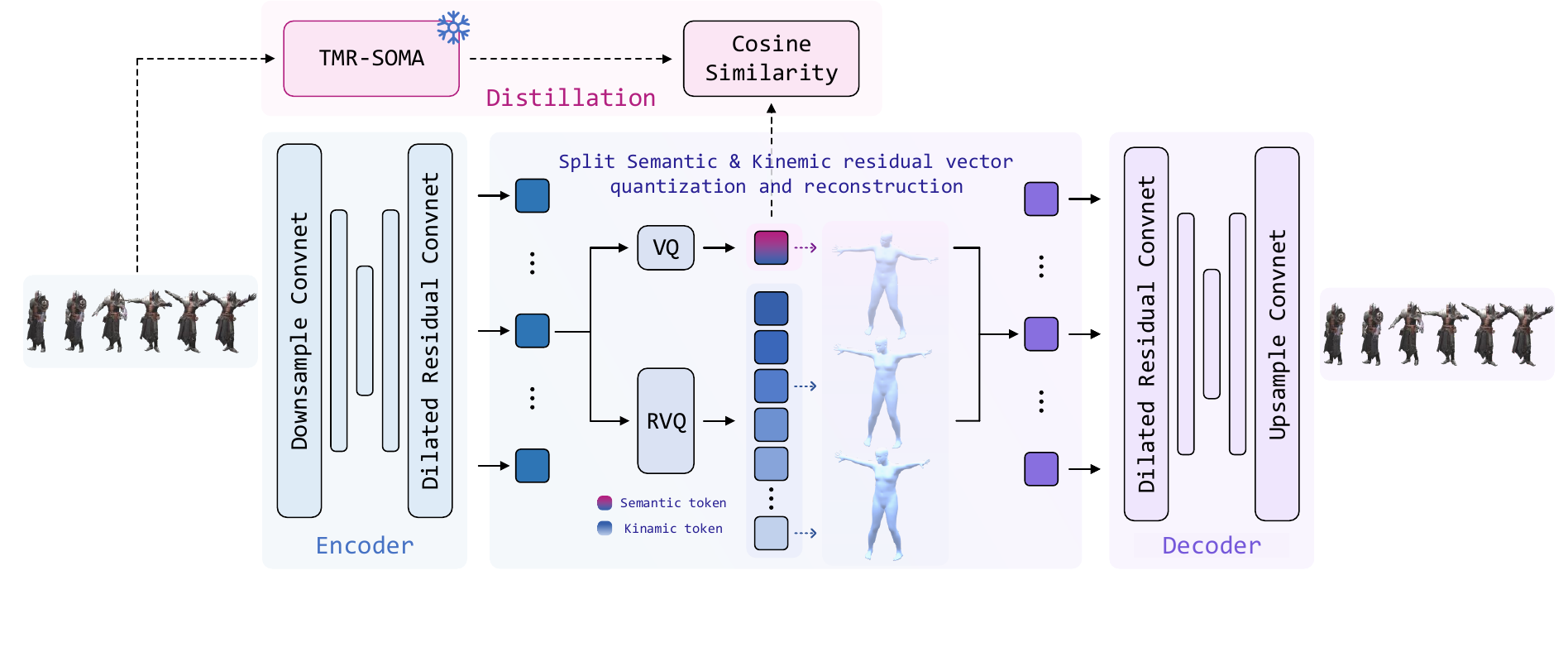}
  \caption{\textbf{SeMoCo tokenizer.} Given an input motion sequence, the
  encoder produces a continuous latent that is factorized into a semantic
  stream and a kinematic residual stream. The semantic stream is
  vector-quantized and aligned with the TMR-SOMA embedding space, whereas RVQ
  progressively quantizes the residual stream to preserve kinematic detail.
  Their quantized outputs are summed at the decoder input to reconstruct the
  motion; semantic supervision is used only during codec training.}
  \label{fig:semoco-tokenizer}
\end{figure*}

\subsection{Semantic-First Motion Codec}
\label{sec:semoco}

SeMoCo converts the 50-Hz transition sequence into a lower-rate sequence of
discrete motion packets. A temporal encoder compresses the input by a factor
of four,
\begin{equation}
  \mathbf{h}_{1:T}
  = E\!\left(\mathbf{u}_{1:N-1}\right),
  \qquad T \approx N/4,
  \label{eq:motion-encoding}
\end{equation}
where each $\mathbf{h}_t$ summarizes a short motion interval at 12.5 Hz. SeMoCo
assigns complementary roles to the codes describing this interval: a primary
code is aligned with motion semantics, while a residual hierarchy preserves
the kinematic information required for reconstruction.

\paragraph{Split semantic--kinematic quantization.}
A conventional residual vector quantizer organizes its codebooks according to
successive reconstruction errors. Its first code also determines the residual
processed by all later stages, so directly imposing semantic supervision on
that code couples motion--language alignment with the geometry of the entire
residual chain. SeMoCo instead separates the two roles into parallel
quantization paths: a single vector quantizer for semantic information and an
independent residual vector quantizer for kinematic reconstruction.

For each encoded interval, the two branches form separate projections,
\begin{equation}
  \mathbf{s}_t=P_{\mathrm{sem}}^{\mathrm{in}}(\mathbf{h}_t),
  \qquad
  \mathbf{k}_t=P_{\mathrm{kin}}^{\mathrm{in}}(\mathbf{h}_t).
  \label{eq:semoco-projections}
\end{equation}
The semantic projection is discretized by a single codebook,
\begin{equation}
  q_t^{\mathrm{sem}}
  = \arg\min_j
  \left\|
    \mathbf{s}_t-\mathbf{e}^{\mathrm{sem}}_j
  \right\|_2^2,
  \label{eq:semantic-quantization}
\end{equation}
while an $L$-stage RVQ successively quantizes the kinematic projection:
\begin{align}
  \mathbf{r}_t^{0}
  &=\mathbf{k}_t,\qquad
  q_t^{\mathrm{kin},\ell}
  =\operatorname*{arg\,min}_{j}
  \left\|
    \mathbf{r}_t^{\ell-1}
    -\mathbf{e}^{\mathrm{kin},\ell}_{j}
  \right\|_2^2,
  \nonumber\\
  \mathbf{r}_t^{\ell}
  &=\mathbf{r}_t^{\ell-1}
  -\mathbf{e}^{\mathrm{kin},\ell}_{q_t^{\mathrm{kin},\ell}},
  \qquad \ell=1,\ldots,L.
  \label{eq:kinematic-rvq}
\end{align}

The quantized outputs of the two branches are then mapped into a common
decoder space and fused additively for reconstruction:
\begin{align}
  \mathbf{z}^{\mathrm{sem}}_t
  &=
  P_{\mathrm{sem}}^{\mathrm{out}}
  \!\left(
    \mathbf{e}^{\mathrm{sem}}_{
      q_t^{\mathrm{sem}}
    }
  \right),
  \nonumber\\
  \mathbf{z}^{\mathrm{kin}}_t
  &=
  P_{\mathrm{kin}}^{\mathrm{out}}
  \!\left(
    \sum_{\ell=1}^{L}
    \mathbf{e}^{\mathrm{kin},\ell}_{
      q_t^{\mathrm{kin},\ell}
    }
  \right),
  \nonumber\\
  \widehat{\mathbf{u}}_{1:N-1}
  &=
  D\!\left(
    \mathbf{z}^{\mathrm{sem}}_{1:T}
    +
    \mathbf{z}^{\mathrm{kin}}_{1:T}
  \right).
  \label{eq:split-quantization}
\end{align}
Accordingly, each interval is represented by a semantic-first packet
\begin{equation}
  \mathbf{m}_t
  =\left[
    q_t^{\mathrm{sem}},
    q_t^{\mathrm{kin},1},\ldots,
    q_t^{\mathrm{kin},L}
  \right].
  \label{eq:motion-packet}
\end{equation}
All codes in $\mathbf{m}_t$ describe the same temporally downsampled interval.
The semantic and kinematic branches are not required to be
information-exclusive; the split instead provides them with distinct
supervision and quantization paths while retaining a shared reconstruction
space.

\paragraph{Window-level semantic distillation.}
Motion semantics are expressed by short-term temporal evolution rather than an
isolated pose. We therefore supervise the primary code stream at the window
level. For a temporal training window $\mathcal{W}$, let
$\mathbf{z}^{\mathrm{sem}}_{\mathcal{W}}$ denote its sequence of quantized
semantic embeddings, and let $\Phi_{\mathrm{TMR}}$ be the frozen motion
encoder from TMR~\cite{petrovich2023tmr}. A training-only temporal head $G$
aggregates the semantic sequence and is aligned with the teacher embedding by
the cosine-distance loss
\begin{equation}
  \mathcal{L}_{\mathrm{sem}}
  = 1 - \operatorname{cos}\!\left(
      G(\mathbf{z}^{\mathrm{sem}}_{\mathcal{W}}),
      \operatorname{sg}\!\left[
        \Phi_{\mathrm{TMR}}(\mathbf{x}_{\mathcal{W}})
      \right]
    \right),
  \label{eq:semantic-alignment}
\end{equation}
where $\operatorname{cos}(\cdot,\cdot)$ denotes cosine similarity and
$\operatorname{sg}[\cdot]$ denotes stop-gradient. This loss injects
window-level action information into the primary stream, while the independent
kinematic RVQ remains organized by reconstruction residuals.

\paragraph{Reconstruction objective.}
Let $\mathbf{P}$ and $\widehat{\mathbf{P}}$ collect the 3D joint positions
obtained by applying the recovery map of Eq.~\eqref{eq:umr-materialization} to
the input and reconstructed transition records, respectively. To preserve
motion geometry and local dynamics, we use
\begin{align}
  \mathcal{L}_{\mathrm{rec}}
  ={}&\mathcal{L}_{\mathrm{pos}}
  +\lambda_{\mathrm{vel}}\mathcal{L}_{\mathrm{vel}}
  +\lambda_{\mathrm{acc}}\mathcal{L}_{\mathrm{acc}}
  \nonumber\\
  &+\lambda_{\mathrm{skate}}\mathcal{L}_{\mathrm{skate}}
  +\lambda_{\mathrm{VQ}}\mathcal{L}_{\mathrm{VQ}},
  \label{eq:reconstruction-objective}
\end{align}
where $\mathcal{L}_{\mathrm{pos}}$ is the $\ell_1$ loss between $\mathbf{P}$
and $\widehat{\mathbf{P}}$, and $\mathcal{L}_{\mathrm{vel}}$ and
$\mathcal{L}_{\mathrm{acc}}$ apply the same loss to their first- and
second-order temporal differences, respectively.
$\mathcal{L}_{\mathrm{skate}}$ penalizes horizontal foot velocity at
ground-truth contact frames, and $\mathcal{L}_{\mathrm{VQ}}$ is the usual
codebook and commitment loss for both quantization branches. All $\lambda$'s
are scalar loss weights. Architecture and optimization details are provided in
the appendix (\emph{SeMoCo Architecture and Training Details}).


\subsection{Dual-Axis Motion Language Model}
\label{sec:dual-axis-transformer}

The semantic-first packets produced by SeMoCo expose two complementary
dependencies for motion modeling. Across time, the model must capture the
evolution of semantic motion states; within each packet, it must resolve the
kinematic details associated with the current state. We model these
dependencies with a temporal Transformer and a lightweight depth decoder.
The temporal Transformer predicts the semantic code of the next packet from
the preceding packet history, while the depth decoder subsequently generates
its residual kinematic codes. This factorization keeps temporal progression
separate from within-packet refinement, rather than serializing both into a
single flattened token stream.

\begin{figure}[!t]
  \centering
  \providecommand{\mtfigwidth}{1}
  \includegraphics[width=\mtfigwidth\linewidth]{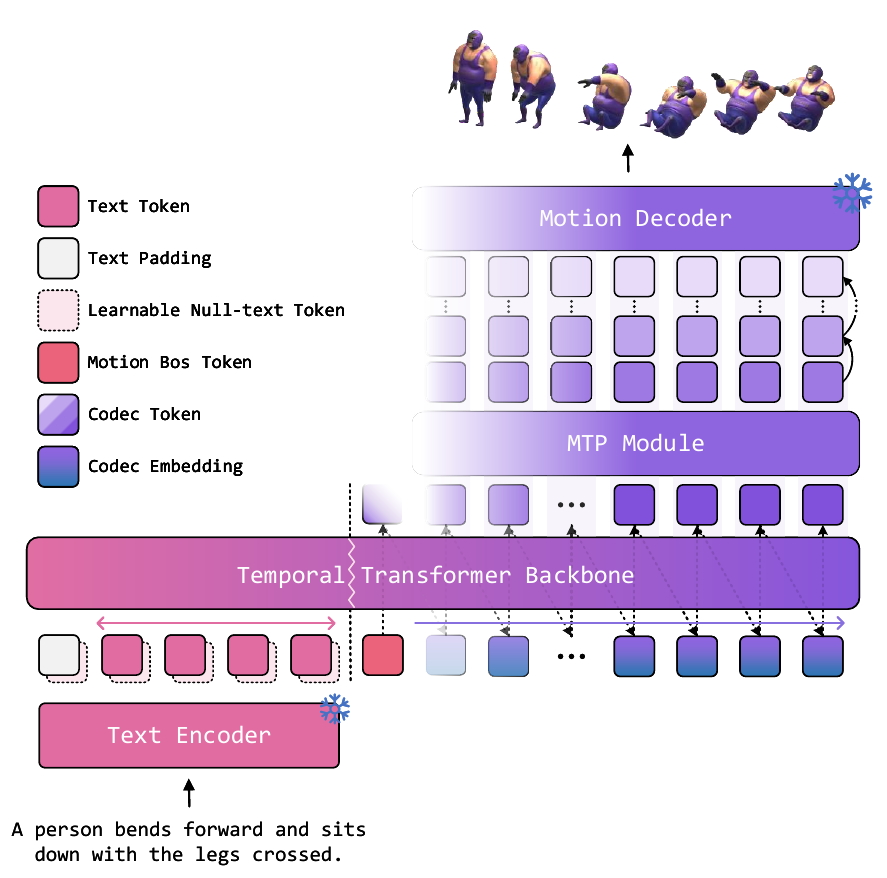}
  \caption{\textbf{Dual-axis motion language model.}
  The temporal Transformer models language-conditioned motion packets across
  time, while the MTP module completes each packet from its semantic code to
  the kinematic residual codes. The generated packets are decoded into
  full-body motion.}
  \label{fig:dual-axis-transformer}
\end{figure}

\paragraph{Dual-axis packet modeling.}
Each completed packet is mapped to a single temporal token by summing its
codebook-specific embeddings. The temporal Transformer processes the preceding
packet sequence and produces a hidden state for the next motion interval, from
which a semantic head predicts $q_t^{\mathrm{sem}}$. Conditioned on this
temporal state and the predicted semantic code, the depth decoder generates
$q_t^{\mathrm{kin},1:L}$ autoregressively along the codebook axis. Given an
optional external condition $\mathbf{c}$, the resulting distribution
factorizes as
\begin{equation}
\begin{split}
p_{\theta}(\mathbf{m}_{1:T}\mid\mathbf{c})
={}&\prod_{t=1}^{T}
p_{\theta}\!\left(
q_t^{\mathrm{sem}}
\mid \mathbf{m}_{<t},\mathbf{c}
\right)
\\
&\times
\prod_{t=1}^{T}\prod_{\ell=1}^{L}
p_{\theta}\!\left(
q_t^{\mathrm{kin},\ell}
\,\middle|\,
\begin{subarray}{c}
\mathbf{m}_{<t},q_t^{\mathrm{sem}},\\[-0.15em]
q_t^{\mathrm{kin},<\ell},\mathbf{c}
\end{subarray}
\right).
\end{split}
\label{eq:dual-axis-factorization}
\end{equation}

where $q_t^{\mathrm{kin},<\ell}$ denotes the residual codes preceding the
$\ell$-th kinematic codebook. The semantic factor is predicted from the
temporal state, whereas the kinematic factors are produced by the depth
decoder in residual order. Both axes are trained jointly using teacher forcing
and codebook-wise cross-entropy losses.

\paragraph{Task-specific context.}
The same packet factorization supports different motion generation tasks by specifying the context available to the temporal Transformer. For
text-to-motion generation, $\mathbf{c}$ contains token-level representations of the input description, which are placed before the motion sequence and remain accessible throughout causal packet generation. For motion prediction, no external condition is required; instead, the observed packets directly initialize $\mathbf{m}_{<t}$, and the model continues the sequence by
predicting future packets. The generated packets are decoded by SeMoCo into
full-body motion. Additional task-specific architecture and optimization
details are provided in the appendix (\emph{Dual-Axis Motion Transformer Learning}).

\begin{figure}[!t]
  \centering
  \providecommand{\omvfigwidth}{0.78}
  \includegraphics[width=\omvfigwidth\linewidth]{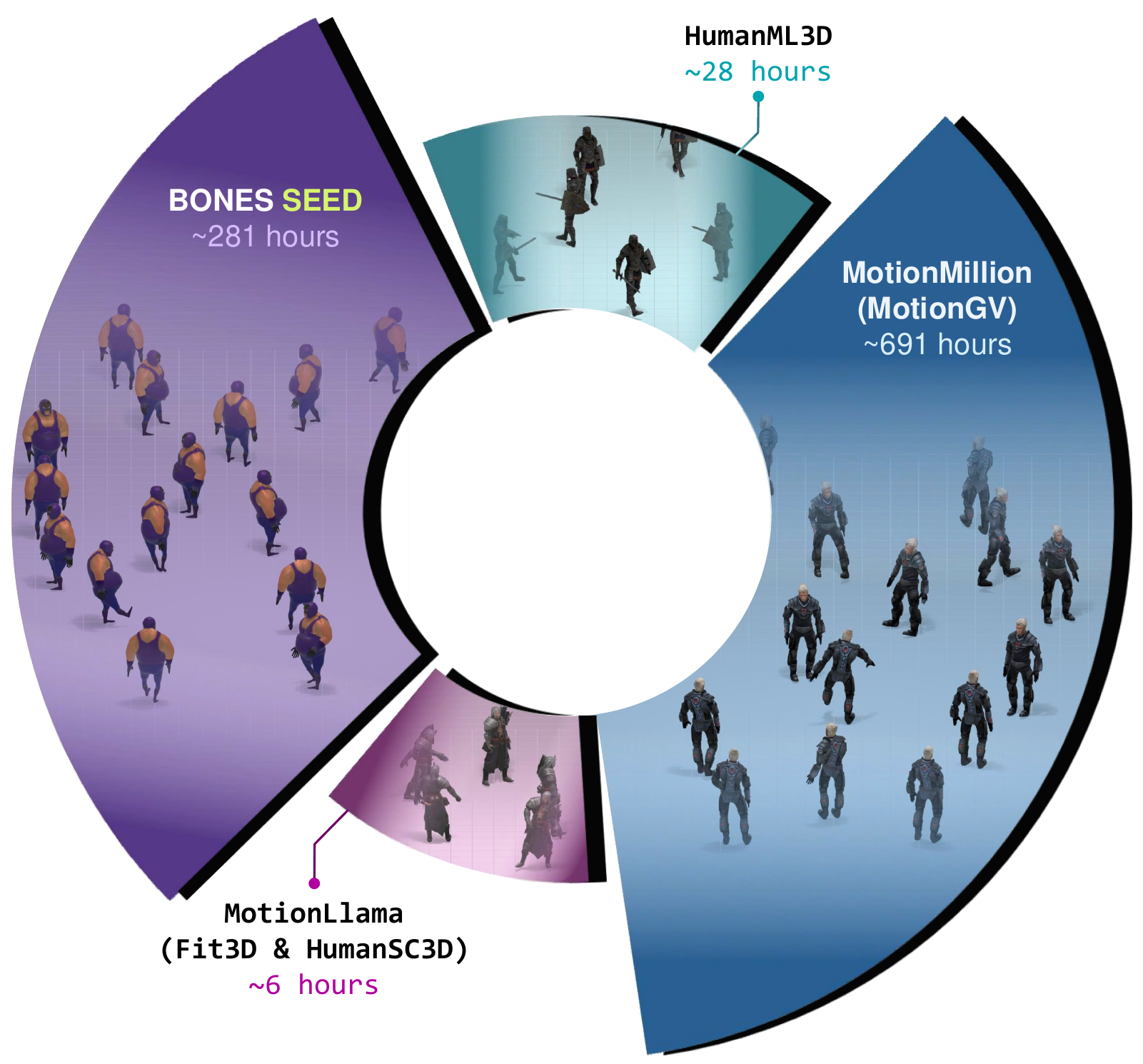}
  \caption{\textbf{$\Omega$-MotionVerse.} Composition of the supporting human motion dataset, comprising around 1,000 hours of text-paired
  motion sequences.}
  \label{fig:omega-motion-verse}
\end{figure}

\section{$\Omega$-MotionVerse}

We construct $\Omega$-MotionVerse by consolidating complementary motion resources
together with their available text annotations into a large-scale corpus of
full-body human motion. The resulting corpus comprises 909,913 text--motion pairs and 1,006 hours of full-body motion, organized into four source groups: MotionGV curated from MotionMillion~\cite{fan2025go}, BONES-SEED~\cite{luo2026sonicsupersizingmotiontracking}, HumanML3D~\cite{guo2022generating}, and a collection of Fit3D~\cite{fieraru2021aifit} and HumanSC3D~\cite{fieraru2021learning} motions curated from MotionHub~\cite{ling2025versatilemotionunifiedframeworkmotion}. These source groups span monocular-video reconstructions and marker-based motion capture, encompassing diverse acquisition settings and annotation pipelines. We retain source labels and annotation
provenance throughout curation, allowing source-wise analysis alongside
aggregate evaluation. Figure~\ref{fig:omega-motion-verse} summarizes the four source groups and their respective durations.

To reconcile heterogeneous conventions, we convert all motions to the SOMA
skeleton~\cite{saito2026somaunifyingparametrichuman}, resample them to
50\,Hz, and apply a shared floor-aligned canonicalization. The resulting
motions are encoded using the previously defined  representation, placing the corpus in a common geometric
and temporal space for joint training.

We construct recording-level splits to prevent related segments from appearing
across partitions. Duplicate motions are removed through content-hash
deduplication, while multiple captions associated with the same HumanML3D
motion are retained as distinct text--motion pairs. Additional corpus
statistics, source-specific conversion procedures, and split details are
provided in the appendix
(\emph{$\Omega$-MotionVerse Composition and Splits}).

\section{Experiments}


\subsection{Motion Reconstruction.}

We evaluate the reconstruction fidelity of each motion tokenizer by encoding and decoding each test sequence without text conditioning, with each method operating in its native motion representation. Joint positions follow the standard HumanML3D 22 SMPL joint convention, and all errors are reported in millimeters. MPJPE measures the mean per-joint position error after pelvis alignment; Med.\ denotes the median sequence-level MPJPE, and PA-MPJPE further applies a per-sequence similarity alignment. The baselines are evaluated on the official HumanML3D test set, while SeMoCo is evaluated on the HumanML3D portion of our test split, which preserves the original HumanML3D train/test partition.
\begin{table}[htbp]
\centering
\small
\setlength{\tabcolsep}{1mm}
\adjustbox{max width=\linewidth}{%
\begin{tabular}{@{}lrrr@{}}
\toprule
Method & MPJPE$\downarrow$ & Med.$\downarrow$ & PA-MPJPE$\downarrow$ \\
\midrule
MoMask & \underline{32.39} & \underline{25.54} & 20.14 \\
MotionGPT3 & 42.38 & 31.58 & 33.49 \\
MotionMillion & 42.54 & 31.51 & \underline{19.84} \\
\midrule
\textbf{Ours} & \textbf{19.22} & \textbf{17.36} & \textbf{16.42} \\
\bottomrule
\end{tabular}}%
\caption{Motion reconstruction results on HumanML3D.}
\label{tab:recon_home}
\end{table}

Table~\ref{tab:recon_home} shows that SeMoCo reconstructs motion more
accurately than MoMask~\cite{guo2024momask},
MotionGPT3~\cite{zhu2025motiongpt3}, and
MotionMillion~\cite{fan2025go} across all metrics.

\subsection{Model variants.}
We evaluate two capacities of the dual-axis motion language model:
Ours-Lite with approximately 150M parameters and Ours-Base with approximately
400M parameters. Both variants share the same SeMoCo tokenizer, text encoder,
packet factorization, training objective, and evaluation protocol, differing
only in generator capacity.

\subsection{Motion Prediction.}

Motion prediction generates future motion conditioned on an observed motion prefix~\cite{jiang2023motiongpt,wang2026unimotionunifiedframeworkmotiontextvision}. For this task, we train a motion Transformer on top of the same tokenizer. We follow the stochastic prediction protocol of observing $0.5$ s and predicting the subsequent $2$ s~\cite{yuan2020dlow,zhang2021mojo,ma2022multi}. For a single sampled prediction, ADE measures the average joint distance, in meters, between the predicted and ground-truth motions over the predicted segment, while FDE measures the corresponding distance at the final frame.minADE$_{50}$ and minFDE$_{50}$ are obtained by taking the minimum ADE and FDE, respectively, over 50 motion sequences generated from the same observed prefix. All methods are evaluated under this protocol. The baselines are evaluated on the official HumanML3D test set, while SeMoCo is evaluated on the HumanML3D portion of our test split.
\begin{table}[htbp]
\centering
\small
\setlength{\tabcolsep}{1mm}
\adjustbox{max width=\linewidth}{%
\begin{tabular}{@{}lrrrr@{}}
\toprule
Method & ADE$\downarrow$ & FDE$\downarrow$ & minADE$_{50}\downarrow$ &
minFDE$_{50}\downarrow$ \\
\midrule
MotionGPT3 & 2.103 & 3.155 & 1.333 & 1.830 \\
MDM & 2.146 & 3.973 & 0.877 & 1.252 \\
MotionGPT & 2.637 & 3.943 & 1.531 & 2.086 \\
\midrule
\textbf{Ours-Lite} & \textbf{1.228} & \textbf{2.449} & \textbf{0.695} &
  \textbf{1.095} \\
\textbf{Ours-Base} & \underline{1.273} & \underline{2.483} & \underline{0.759} &
  \underline{1.220} \\
\bottomrule
\end{tabular}}%
\caption{Motion prediction results on HumanML3D.}
\label{tab:prediction_hml}
\end{table}

Table~\ref{tab:prediction_hml} shows that Ours-Lite achieves the lowest error across all metrics, while also yielding lower errors than the larger Ours-Base. This reverses the ordering observed for text-to-motion, indicating that the larger model does not provide an advantage for motion prediction under this protocol.

\begin{table*}[t]
\centering
\small
\setlength{\tabcolsep}{1mm}
\adjustbox{max width=\linewidth}{%
\begin{tabular}{@{}lrrrrrrr@{}}
\toprule
Method & FID$\downarrow$ & R@1$\uparrow$ & R@2$\uparrow$ &
R@3$\uparrow$ & R@5$\uparrow$ & MedR$\downarrow$ & Align$\uparrow$ \\
\midrule
MoMask & \textbf{.32\,\textpm\,.09} & \textbf{.452\,\textpm\,.053}
  & .587\,\textpm\,.042 & .656\,\textpm\,.038
  & .734\,\textpm\,.039 & \underline{1.9\,\textpm\,.3} & .957\,\textpm\,.003 \\
HyMotion & \underline{.34\,\textpm\,.08} & \underline{.449\,\textpm\,.054}
  & \textbf{.595\,\textpm\,.040} & \textbf{.669\,\textpm\,.038}
  & \textbf{.747\,\textpm\,.035} & \textbf{1.8\,\textpm\,.4}
  & \textbf{.960\,\textpm\,.004} \\
MotionGPT3 & .39\,\textpm\,.10 & .446\,\textpm\,.052
  & \underline{.590\,\textpm\,.056} & \underline{.660\,\textpm\,.046}
  & \underline{.740\,\textpm\,.038} & \underline{1.9\,\textpm\,.4}
  & \underline{.959\,\textpm\,.004} \\
MotionMillion & 3.01\,\textpm\,.44 & .327\,\textpm\,.047
  & .449\,\textpm\,.053 & .520\,\textpm\,.055
  & .599\,\textpm\,.048 & 3.4\,\textpm\,1.1 & .919\,\textpm\,.005 \\
\midrule
Kimodo$^\ddagger$ & 1.091\,\textpm\,.101 & \textbf{.645\,\textpm\,.122}
  & \textbf{.775\,\textpm\,.128} & \textbf{.873\,\textpm\,.090}
  & \textbf{.923\,\textpm\,.056} & \textbf{1.1\,\textpm\,.4}
  & \textbf{.076\,\textpm\,.011} \\
\textbf{Ours-Lite}$^\ddagger$ & \underline{.920\,\textpm\,.101} & .326\,\textpm\,.103
  & .451\,\textpm\,.084 & .540\,\textpm\,.107
  & .707\,\textpm\,.115 & 3.2\,\textpm\,1.2 & .032\,\textpm\,.009 \\
\textbf{Ours-Base}$^\ddagger$ & \textbf{.913\,\textpm\,.092} & \underline{.422\,\textpm\,.104}
  & \underline{.558\,\textpm\,.095} & \underline{.653\,\textpm\,.086}
  & \underline{.777\,\textpm\,.079} & \underline{2.2\,\textpm\,.8} & \underline{.039\,\textpm\,.009} \\
\bottomrule
\end{tabular}}%
\caption{Text-to-motion evaluation on HumanML3D. Unmarked rows are evaluated
under the standard HumanML3D protocol, while rows marked with $^\ddagger$ are
evaluated in the TMR-SOMA track using native SOMA outputs from the HumanML3D
subset. Both tracks use a retrieval batch size of 32 and report mean $\pm$
standard deviation over 30 repeats. Results are ranked only within the same
evaluation track. Best and second-best results are shown in \textbf{bold} and
\underline{underlined}, respectively.}
\label{tab:t2m_hml}
\end{table*}

\begin{table*}[!htbp]
\centering
\small
\adjustbox{max width=\linewidth}{%
\begin{tabular}{@{}llrrrrr@{}}
\toprule
Subset & Method & FID$\downarrow$ & R@1$\uparrow$ & R@2$\uparrow$ & R@3$\uparrow$ & MedR$\downarrow$ \\
\midrule
\multirow{5}{*}{Overall}
& Kimodo & \textbf{.143\,\textpm\,.008} & \textbf{.553\,\textpm\,.038} & \textbf{.660\,\textpm\,.033} & \textbf{.719\,\textpm\,.036} & \textbf{1.100\,\textpm\,.305} \\
& HyMotion$\dagger$ & .285\,\textpm\,.019 & .302\,\textpm\,.028 & .418\,\textpm\,.028 & .484\,\textpm\,.030 & 3.850\,\textpm\,.604 \\
& MotionMillion$\dagger$ & .486\,\textpm\,.028 & .131\,\textpm\,.020 & .190\,\textpm\,.026 & .230\,\textpm\,.017 & 25.783\,\textpm\,4.527 \\
\cmidrule(lr){2-7}
& \textbf{Ours-Lite} & .186\,\textpm\,.010 & .484\,\textpm\,.031 & .600\,\textpm\,.034 & .656\,\textpm\,.039 & 1.700\,\textpm\,.447 \\
& \textbf{Ours-Base} & \underline{.181\,\textpm\,.010} & \underline{.494\,\textpm\,.039} & \underline{.614\,\textpm\,.031} & \underline{.670\,\textpm\,.029} & \underline{1.500\,\textpm\,.491} \\
\midrule
\multirow{5}{*}{HumanML3D}
& Kimodo & 1.091\,\textpm\,.101 & \underline{.645\,\textpm\,.122} & \underline{.775\,\textpm\,.128} & \underline{.873\,\textpm\,.090} & \underline{1.133\,\textpm\,.434} \\
& HyMotion$\dagger$ & \textbf{.645\,\textpm\,.087} & \textbf{.647\,\textpm\,.130} & \textbf{.811\,\textpm\,.117} & \textbf{.883\,\textpm\,.087} & \textbf{1.083\,\textpm\,.265} \\
& MotionMillion$\dagger$ & .963\,\textpm\,.099 & .386\,\textpm\,.114 & .515\,\textpm\,.125 & .579\,\textpm\,.122 & 2.817\,\textpm\,1.329 \\
\cmidrule(lr){2-7}
& \textbf{Ours-Lite} & .920\,\textpm\,.101 & .326\,\textpm\,.103 & .451\,\textpm\,.084 & .540\,\textpm\,.107 & 3.233\,\textpm\,1.158 \\
& \textbf{Ours-Base} & \underline{.913\,\textpm\,.092} & .422\,\textpm\,.104 & .558\,\textpm\,.095 & .653\,\textpm\,.086 & 2.167\,\textpm\,.834 \\
\midrule
\multirow{5}{*}{bones-seed}
& Kimodo & \textbf{.100\,\textpm\,.009} & \textbf{.663\,\textpm\,.049} & \textbf{.777\,\textpm\,.039} & \textbf{.820\,\textpm\,.037} & \textbf{1.000\,\textpm\,.000} \\
& HyMotion$\dagger$ & .399\,\textpm\,.023 & .297\,\textpm\,.033 & .403\,\textpm\,.039 & .465\,\textpm\,.039 & 4.217\,\textpm\,.944 \\
& MotionMillion$\dagger$ & .621\,\textpm\,.037 & .072\,\textpm\,.025 & .127\,\textpm\,.029 & .162\,\textpm\,.029 & 35.983\,\textpm\,6.655 \\
\cmidrule(lr){2-7}
& \textbf{Ours-Lite} & .219\,\textpm\,.016 & .518\,\textpm\,.039 & .654\,\textpm\,.035 & .715\,\textpm\,.045 & 1.350\,\textpm\,.476 \\
& \textbf{Ours-Base} & \underline{.202\,\textpm\,.014} & \underline{.539\,\textpm\,.048} & \underline{.671\,\textpm\,.044} & \underline{.733\,\textpm\,.043} & \underline{1.167\,\textpm\,.379} \\
\midrule
\multirow{5}{*}{MotionGV}
& Kimodo & 1.001\,\textpm\,.097 & .308\,\textpm\,.103 & .440\,\textpm\,.123 & .532\,\textpm\,.130 & 3.283\,\textpm\,1.388 \\
& HyMotion$\dagger$ & .895\,\textpm\,.134 & .526\,\textpm\,.133 & .683\,\textpm\,.090 & .771\,\textpm\,.090 & 1.433\,\textpm\,.504 \\
& MotionMillion$\dagger$ & \textbf{.795\,\textpm\,.093} & .517\,\textpm\,.105 & .676\,\textpm\,.103 & .758\,\textpm\,.097 & 1.517\,\textpm\,.594 \\
\cmidrule(lr){2-7}
& \textbf{Ours-Lite} & \underline{.805\,\textpm\,.094} & \underline{.712\,\textpm\,.104} & \underline{.854\,\textpm\,.085} & \underline{.924\,\textpm\,.054} & \textbf{1.000\,\textpm\,.000} \\
& \textbf{Ours-Base} & .812\,\textpm\,.098 & \textbf{.724\,\textpm\,.111} & \textbf{.883\,\textpm\,.082} & \textbf{.932\,\textpm\,.056} & \underline{1.067\,\textpm\,.254} \\
\bottomrule
\end{tabular}}%
\caption{TMR-SOMA text-to-motion retrieval by provenance (256 clips per
repeat; mean $\pm$ standard deviation over 30 repeats). HyMotion and
MotionMillion ($\dagger$) are evaluated after SMPL-to-SOMA retargeting.}
\label{tab:t2m_soma_retrieval}
\end{table*}

\subsection{Text-to-motion.}

\paragraph{Evaluator Spaces.} 
We evaluate text-to-motion in two separate tracks defined by the native motion representation of each method. Methods producing motion in the SMPL or HumanML3D convention are
evaluated under the standard HumanML3D protocol using its official evaluator~\cite{guo2022generating}. Methods producing full-body SOMA motion are instead evaluated with a TMR-based motion--text evaluator operating directly on the SOMA representation, which we refer to as TMR-SOMA~\cite{petrovich2023tmr}. The evaluator is initialized from the model released with Kimodo~\cite{rempe2026kimodo}, fine-tuned only on the $\Omega$-MotionVerse training partition, and then frozen for evaluation. Metric values are compared only within the same evaluation track, since the evaluated models produce outputs in different native motion representations.

\begin{figure*}[!htbp]
  \centering
  \IfFileExists{figures/images/t2m_qualitative.pdf}{%
    \includegraphics[width=\textwidth]{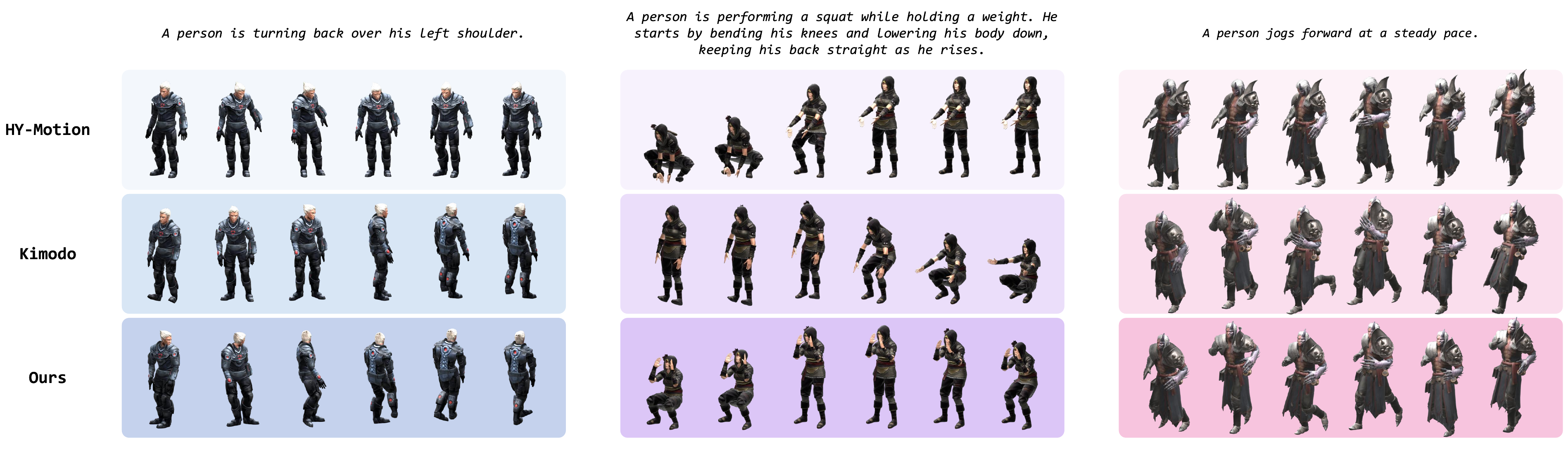}%
  }{%
    \setlength{\fboxsep}{0pt}%
    \fbox{\parbox[c][0.14\textheight][c]{0.985\textwidth}{%
      \centering
      \begin{tabular}{cccc}
        \textbf{Ground Truth} & \textbf{HyMotion} & \textbf{Kimodo} &
        \textbf{Ours-Base} \\
        \multicolumn{4}{c}{\rule{0pt}{34pt}\textit{Example A: ordered action composition}} \\
        \multicolumn{4}{c}{\rule{0pt}{34pt}\textit{Example B: directional or body-part constraint}}
      \end{tabular}
    }}%
  }
  \caption{\textbf{Qualitative text-to-motion comparison.}
Generated motion sequences from HY-Motion, Kimodo, and Ours-Base for three representative text prompts, visualized at matched timestamps.}
  \label{fig:t2m-qualitative}
\end{figure*}

\begin{table*}[!htbp]
\centering
\small
\setlength{\tabcolsep}{1mm}
\adjustbox{max width=\linewidth}{%
\begin{tabular}{@{}llrrrrrrr@{}}
\toprule
RVQ layout & Sem. & FID$\downarrow$ & R@1$\uparrow$ & R@2$\uparrow$ &
R@3$\uparrow$ & R@5$\uparrow$ & MedR$\downarrow$ & MPJPE-77$\downarrow$ \\
\midrule
Single-chain & $\times$ & \underline{.202\,\textpm\,.013}
  & \underline{.483\,\textpm\,.034} & \textbf{.607\,\textpm\,.036}
  & \textbf{.663\,\textpm\,.034} & \textbf{.722\,\textpm\,.032}
  & \textbf{1.65\,\textpm\,.48} & \underline{15.60} \\
Single-chain & $\checkmark$ & .211\,\textpm\,.014
  & .451\,\textpm\,.031 & .579\,\textpm\,.024
  & .637\,\textpm\,.027 & \underline{.719\,\textpm\,.035}
  & 1.93\,\textpm\,.25 & 18.54 \\
Split-branch & $\times$ & .226\,\textpm\,.013
  & .319\,\textpm\,.033 & .425\,\textpm\,.037
  & .490\,\textpm\,.039 & .561\,\textpm\,.037
  & 3.83\,\textpm\,.93 & \textbf{13.70} \\
\textbf{Split-branch} & $\checkmark$ & \textbf{.186\,\textpm\,.010}
  & \textbf{.484\,\textpm\,.031} & \underline{.600\,\textpm\,.034}
  & \underline{.656\,\textpm\,.039} & .717\,\textpm\,.038
  & \underline{1.70\,\textpm\,.45} & 15.93 \\
\bottomrule
\end{tabular}}%
\caption{Tokenizer ablation with matched 150M Flan-T5 generators. TMR-SOMA
Overall values are mean $\pm$ standard deviation over 30 repeats; generation
pools are not clip-paired. MPJPE-77 uses the shared reconstruction split (mm).}
\label{tab:ab_t2m_tokenizer}
\end{table*}

\paragraph{Evaluation Metrics.}
Following common text-to-motion evaluation practice~\cite{guo2024momask},
we assess distributional fidelity and text--motion correspondence within each
evaluator space. FID is the Fr\'echet distance between generated and reference
motion-feature distributions, with lower values indicating a closer match.
R@k is the fraction of captions whose paired motion is retrieved within the
top $k$ candidates, whereas MedR is the median rank of that paired motion;
higher R@k and lower MedR are better. Align, where reported, is the mean
cosine similarity of paired text and motion embeddings, so higher values
indicate stronger correspondence. All metrics are computed with the
corresponding frozen evaluator and are compared only within that evaluator
space.

Table~\ref{tab:t2m_hml} summarizes representative text-to-motion results in
the two evaluator spaces, with rows evaluated by TMR-SOMA marked by
$^\ddagger$. Within the TMR-SOMA track, Ours-Base improves R@1 from $.326$ to
$.422$ and reduces FID from $.920$ to $.913$ compared with
Ours-Lite, while Kimodo remains stronger in retrieval. Table~\ref{tab:t2m_soma_retrieval} reports TMR-SOMA results on the complete
test pool and on the HumanML3D, bones-seed, and MotionGV subsets.

Ours-Base consistently outperforms Ours-Lite, whereas
Kimodo achieves the strongest overall performance. The relative ordering,
however, varies across sources: Ours-Base outperforms all the other baselines on MotionGV, Kimodo performs best on bones-seed, and HyMotion
leads on the HumanML3D subset. Additional distribution, alignment, and motion-quality metrics are reported in appendix (\emph{Full TMR-SOMA Motion Metrics}).

\subsection{Qualitative Comparison}

Figure~\ref{fig:t2m-qualitative} presents three representative comparisons of language-conditioned motion generation. Our model produces semantically consistent and temporally coherent motions across diverse text descriptions, providing qualitative evidence of its effectiveness.

\subsection{Ablation Studies}
\paragraph{Semantic--geometry trade-off.}
Table~\ref{tab:ab_t2m_tokenizer} varies the RVQ layout and semantic
distillation objective under matched generator capacity, reporting TMR-SOMA
generation together with native SOMA reconstruction.

Split-branch RVQ with semantic supervision obtains the lowest FID ($.186$) and
highest mean R@1 ($.484$), while Plain RVQ remains strongest on R@2--R@5 and
median rank. Within the split layout, semantic supervision changes FID from
$.226$ to $.186$ and R@1 from $.319$ to $.484$, while MPJPE-77 rises from
$13.70$ to $15.93$ mm. The Single-chain contrast does not show the same
generation gain, bounding the result to the split semantic configuration.
Source-wise T2M results and full reconstruction diagnostics appear in  
the appendix (\emph{Source-wise Tokenizer Ablation} and
\emph{Full Tokenizer Reconstruction Ablation}).


\section{Conclusion}
SeMoCo couples a teacher-aligned semantic code and kinematic residual hierarchy
with a dual-axis packet generator. Experiments on $\Omega$-MotionVerse span
text-to-motion, reconstruction, and prediction, revealing a reconstruction
cost for semantic supervision and task-dependent effects of model scale. These
results support semantic packet structure without implying complete
semantic--kinematic separation; HML-263 and TMR-SOMA remain distinct evaluator
spaces.

\bibliographystyle{plainnat}
\bibliography{references}

\clearpage
\beginappendix

\section{Additional Experimental Results}
\label{sup:additional-results}

\subsection{Shared-Split Reconstruction}
\label{sup:recon-shared}

Table~\ref{tab:recon_shared} evaluates every method on the same test split.
Each external method is converted into and out of its native representation,
and all errors are measured in a common pelvis-aligned output space over the
22 SMPL body joints.
\begin{table}[htbp]
\centering
\small
\setlength{\tabcolsep}{1mm}
\adjustbox{max width=\linewidth}{%
\begin{tabular}{@{}lrrr@{}}
\toprule
Method & MPJPE-22$\downarrow$ & Med.$\downarrow$ & PA-MPJPE$\downarrow$ \\
\midrule
MoMask & 90.02 & 75.92 & 63.39 \\
MotionGPT3 & 96.01 & 85.40 & 77.19 \\
MotionMillion & \underline{78.95} & \underline{68.00} & \underline{53.43} \\
\midrule
\textbf{Ours} & \textbf{12.83} & \textbf{10.13} & \textbf{11.07} \\
\bottomrule
\end{tabular}}%
\caption{Shared-split reconstruction results.}
\label{tab:recon_shared}
\end{table}

Our MPJPE-22 is 12.83\,mm, against 78.95\,mm for MotionMillion, 90.02\,mm for
MoMask, and 96.01\,mm for MotionGPT3.

\subsection{Reconstruction Diagnostics}
\label{sup:recon-diagnostics}

\begin{figure*}[!t]
  \centering
  \includegraphics[width=\textwidth]{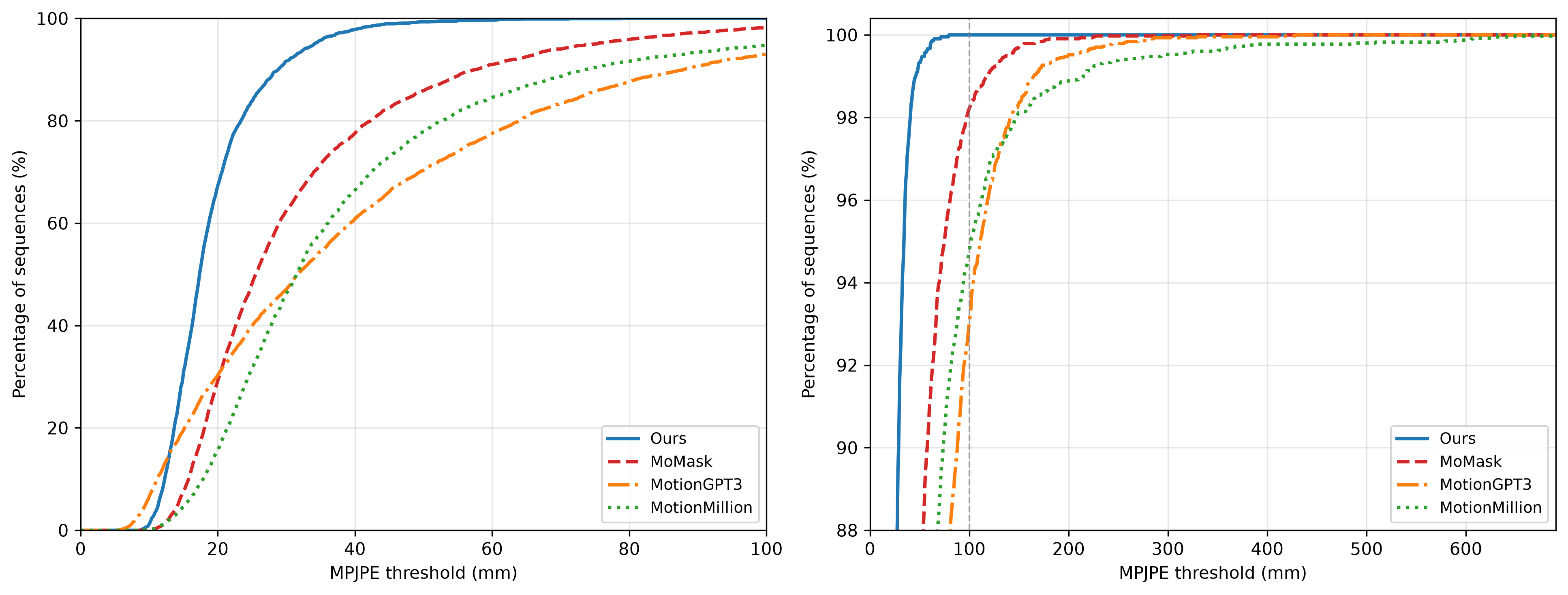}
  \caption{\textbf{Per-sequence reconstruction error.} Cumulative distribution
  of MPJPE, with each method evaluated on its own test set: ours on the
  HumanML3D portion of our split ($n=2{,}108$), MoMask and MotionGPT3 on the
  official HumanML3D test set ($n=4{,}372$), and MotionMillion on its own test
  set ($n=4{,}042$). \emph{Left:} 0--100\,mm. \emph{Right:} the upper 12\% of
  each distribution, where the dashed line marks the range of the left panel.}
  \label{fig:recon-cdf}
\end{figure*}

Figure~\ref{fig:recon-cdf} resolves the reconstruction averages of the main
paper into per-sequence distributions. Our mean and median errors are 19.2 and
17.4\,mm, and the worst single sequence reaches 79\,mm, against 324, 428, and
691\,mm for MoMask, MotionGPT3, and MotionMillion.

\begin{figure*}[!t]
  \centering
  \includegraphics[width=\textwidth]{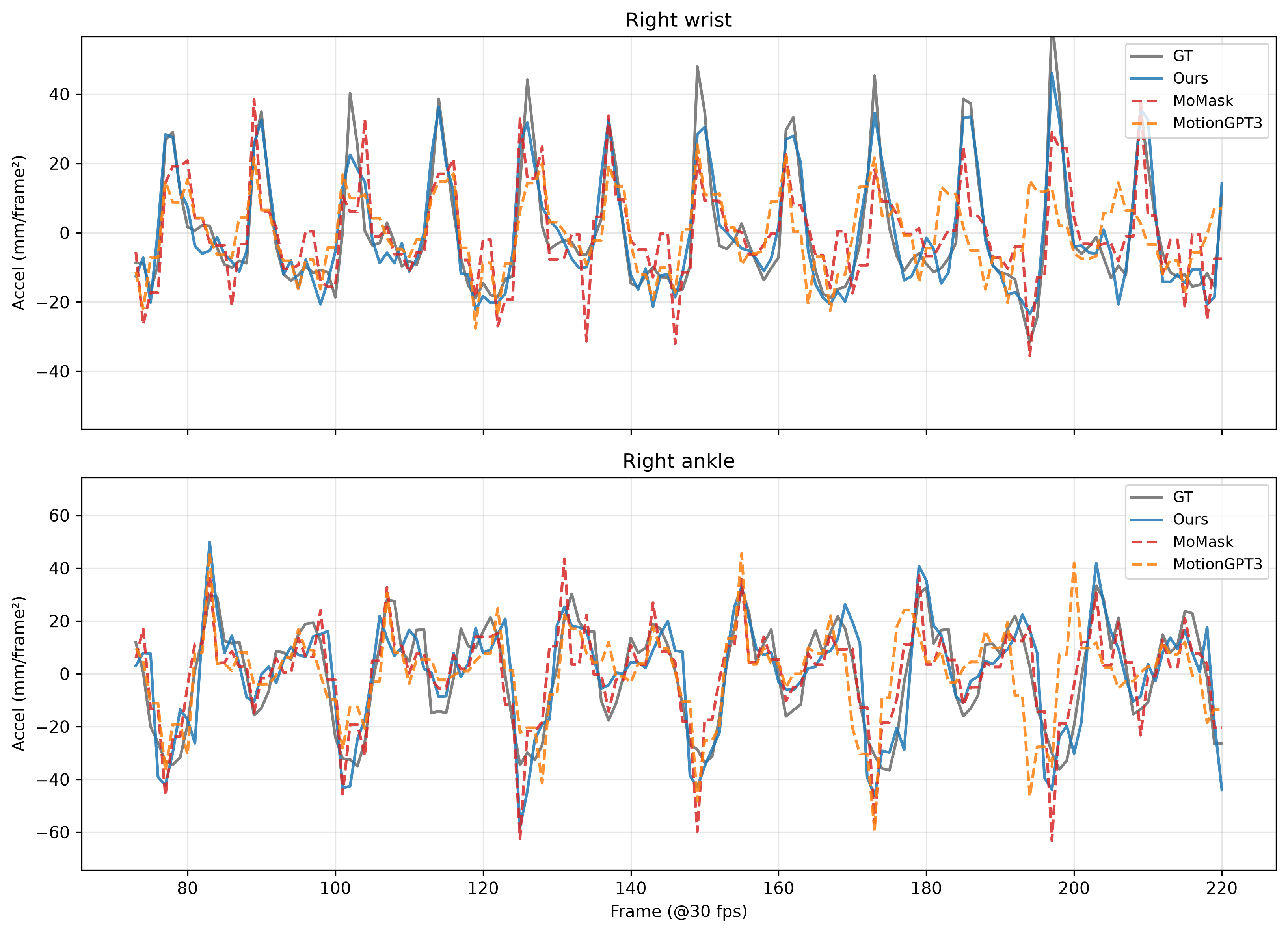}
  \caption{\textbf{Per-frame vertical acceleration} of the right wrist and
  right ankle on the most dynamic clip of our test set, shown over its busiest
  150-frame window with all methods resampled to 30\,fps and pelvis aligned.
  MotionMillion is not included.}
  \label{fig:recon-jitter}
\end{figure*}

Figure~\ref{fig:recon-jitter} inspects temporal smoothness on one clip. At the
right wrist our acceleration standard deviation is 10.4 against a ground-truth
12.1, whereas MotionGPT3 flattens the signal to 7.5 and MoMask matches its
scale at 12.6 but with a larger frame-wise error. The corresponding RMSE values
are 9.3, 22.7, and 18.7\,mm/frame$^2$.

\subsection{Full TMR-SOMA Motion Metrics}
\label{sup:soma-motion-metrics}

We complement the retrieval study in the main paper with distribution and
motion-quality diagnostics. Table~\ref{tab:t2m_soma_motion} reports MM-Dist,
alignment, diversity, foot skating, and jerk for each dataset provenance.
\begin{table*}[t]
\centering
\adjustbox{max width=\textwidth}{%
\begin{tabular}{llrrrrr}
\toprule
Subset & Method & MM-Dist$\downarrow$ & Align$\uparrow$ & Diversity$\uparrow$ & FootSkate$\downarrow$ & Jerk$\downarrow$ \\
\midrule
\multirow{5}{*}{Overall}
& Kimodo & \textbf{1.343\,\textpm\,.002} & \textbf{.097\,\textpm\,.003} & \textbf{1.385\,\textpm\,.007} & \textbf{.024\,\textpm\,.001} & \textbf{73.329\,\textpm\,6.589} \\
& HyMotion$\dagger$ & 1.376\,\textpm\,.001 & .052\,\textpm\,.002 & 1.348\,\textpm\,.008 & \underline{.047\,\textpm\,.003} & \underline{115.424\,\textpm\,19.185} \\
& MotionMillion$\dagger$ & 1.404\,\textpm\,.002 & .013\,\textpm\,.002 & 1.264\,\textpm\,.010 & .062\,\textpm\,.002 & 438.251\,\textpm\,22.803 \\
\cmidrule(lr){2-7}
& \textbf{Ours-Lite} & 1.357\,\textpm\,.002 & .078\,\textpm\,.003 & 1.351\,\textpm\,.008 & .068\,\textpm\,.002 & 406.691\,\textpm\,32.168 \\
& \textbf{Ours-Base} & \underline{1.355\,\textpm\,.002} & \underline{.081\,\textpm\,.002} & \underline{1.355\,\textpm\,.007} & .068\,\textpm\,.001 & 432.748\,\textpm\,23.579 \\
\midrule
\multirow{5}{*}{HumanML3D}
& Kimodo & \textbf{1.359\,\textpm\,.008} & \textbf{.076\,\textpm\,.011} & \textbf{1.382\,\textpm\,.014} & \textbf{.028\,\textpm\,.004} & \underline{81.068\,\textpm\,24.207} \\
& HyMotion$\dagger$ & \textbf{1.359\,\textpm\,.008} & \textbf{.076\,\textpm\,.010} & \underline{1.293\,\textpm\,.031} & \underline{.038\,\textpm\,.004} & \textbf{74.812\,\textpm\,19.972} \\
& MotionMillion$\dagger$ & 1.395\,\textpm\,.008 & .027\,\textpm\,.012 & 1.185\,\textpm\,.037 & \textbf{.028\,\textpm\,.006} & 293.387\,\textpm\,45.393 \\
\cmidrule(lr){2-7}
& \textbf{Ours-Lite} & 1.391\,\textpm\,.007 & .032\,\textpm\,.009 & 1.239\,\textpm\,.032 & .078\,\textpm\,.004 & 721.084\,\textpm\,97.224 \\
& \textbf{Ours-Base} & \underline{1.386\,\textpm\,.007} & \underline{.039\,\textpm\,.009} & 1.247\,\textpm\,.035 & .081\,\textpm\,.005 & 874.236\,\textpm\,76.726 \\
\midrule
\multirow{5}{*}{BONES-SEED}
& Kimodo & \textbf{1.334\,\textpm\,.002} & \textbf{.110\,\textpm\,.003} & \textbf{1.376\,\textpm\,.009} & \textbf{.024\,\textpm\,.001} & \textbf{72.185\,\textpm\,8.189} \\
& HyMotion$\dagger$ & 1.381\,\textpm\,.002 & .046\,\textpm\,.002 & 1.334\,\textpm\,.011 & \underline{.048\,\textpm\,.003} & \underline{90.107\,\textpm\,10.635} \\
& MotionMillion$\dagger$ & 1.415\,\textpm\,.002 & -.002\,\textpm\,.003 & 1.256\,\textpm\,.015 & .067\,\textpm\,.002 & 464.235\,\textpm\,29.534 \\
\cmidrule(lr){2-7}
& \textbf{Ours-Lite} & 1.352\,\textpm\,.002 & .085\,\textpm\,.003 & 1.359\,\textpm\,.007 & .067\,\textpm\,.002 & 357.211\,\textpm\,25.063 \\
& \textbf{Ours-Base} & \underline{1.350\,\textpm\,.002} & \underline{.088\,\textpm\,.002} & \underline{1.362\,\textpm\,.008} & .067\,\textpm\,.002 & 348.339\,\textpm\,21.259 \\
\midrule
\multirow{5}{*}{MotionGV}
& Kimodo & 1.387\,\textpm\,.005 & .038\,\textpm\,.007 & \textbf{1.343\,\textpm\,.022} & \textbf{.023\,\textpm\,.004} & \textbf{107.659\,\textpm\,24.945} \\
& HyMotion$\dagger$ & 1.375\,\textpm\,.006 & .054\,\textpm\,.008 & \underline{1.319\,\textpm\,.020} & \underline{.053\,\textpm\,.006} & \underline{405.600\,\textpm\,211.470} \\
& MotionMillion$\dagger$ & 1.372\,\textpm\,.004 & .058\,\textpm\,.006 & 1.265\,\textpm\,.023 & .065\,\textpm\,.006 & 476.107\,\textpm\,112.959 \\
\cmidrule(lr){2-7}
& \textbf{Ours-Lite} & \underline{1.355\,\textpm\,.004} & \underline{.081\,\textpm\,.006} & 1.309\,\textpm\,.020 & .060\,\textpm\,.006 & 506.546\,\textpm\,171.456 \\
& \textbf{Ours-Base} & \textbf{1.354\,\textpm\,.004} & \textbf{.083\,\textpm\,.005} & 1.311\,\textpm\,.026 & .061\,\textpm\,.006 & 512.559\,\textpm\,121.353 \\
\bottomrule
\end{tabular}%
}
\caption{Additional TMR-SOMA metrics corresponding to the retrieval results in
the main paper. Values are mean $\pm$ standard deviation over 30 repeats.
HyMotion and MotionMillion ($\dagger$) are evaluated after SMPL-to-SOMA
retargeting.}
\label{tab:t2m_soma_motion}
\end{table*}

MM-Dist is the mean distance between paired text and motion embeddings, while
Align is their mean raw cosine similarity. Diversity is the mean distance
between randomly sampled generated-motion embeddings. FootSkate and Jerk are
defined with their evaluation units in the protocol section below. HyMotion and
MotionMillion use the same SMPL-to-SOMA retargeting as in the main paper.
Kimodo attains the lowest MM-Dist and the highest Align in the Overall pool
and on BONES-SEED, with Ours-Base second on both; on MotionGV the ordering
reverses and Ours-Base leads both metrics. Kimodo also reports the lowest
FootSkate and jerk on every subset, and our variants reach their highest jerk
on HumanML3D.

\subsection{Text-Encoder Variants}
\label{sup:text-encoder}

We compare Flan-T5-XL~\cite{chung2024scaling},
SigLIP~\cite{zhai2023sigmoid}, and
Qwen3-Embedding~\cite{zhang2025qwen3embeddingadvancingtext} under the
Ours-Lite configuration to test whether their ordering transfers across
evaluator spaces. Table~\ref{tab:ab_t2m_text} reports the two tracks separately.
\begin{table*}[t]
\centering
\small
\setlength{\tabcolsep}{2.8pt}
\adjustbox{max width=\linewidth}{%
\begin{tabular}{@{}lrrrr@{}}
\toprule
\multicolumn{5}{c}{(a) HumanML3D / HML-263 evaluator} \\
\midrule
Text encoder & FID$\downarrow$ & R@1$\uparrow$ & R@5$\uparrow$ & MedR$\downarrow$ \\
\midrule
Flan-T5 & 8.75\,\textpm\,1.07 & \textbf{.203\,\textpm\,.037} & \textbf{.442\,\textpm\,.042} & \textbf{7.67\,\textpm\,1.56} \\
SigLIP & \textbf{6.36\,\textpm\,0.69} & .153\,\textpm\,.039 & .382\,\textpm\,.044 & 10.70\,\textpm\,2.93 \\
Qwen3 & \underline{8.15\,\textpm\,1.14} & \underline{.164\,\textpm\,.044} & \underline{.419\,\textpm\,.056} & \underline{8.97\,\textpm\,2.37} \\
\bottomrule
\end{tabular}}%
\par\vspace{2mm}
\adjustbox{max width=\linewidth}{%
\begin{tabular}{@{}lrrrrrr@{}}
\toprule
\multicolumn{7}{c}{(b) TMR-SOMA / overall test set} \\
\midrule
Text encoder & FID$\downarrow$ & R@1$\uparrow$ & R@2$\uparrow$ & R@3$\uparrow$ & MedR$\downarrow$ & Align$\uparrow$ \\
\midrule
Flan-T5 & \textbf{.186\,\textpm\,.010} & \textbf{.484\,\textpm\,.031} & \textbf{.600\,\textpm\,.034} & \textbf{.656\,\textpm\,.039} & \textbf{1.70\,\textpm\,.45} & \textbf{.078\,\textpm\,.003} \\
SigLIP & \underline{.200\,\textpm\,.012} & \underline{.416\,\textpm\,.039} & \underline{.531\,\textpm\,.039} & \underline{.597\,\textpm\,.039} & \underline{2.32\,\textpm\,.53} & \underline{.071\,\textpm\,.003} \\
Qwen3 & .314\,\textpm\,.024 & .135\,\textpm\,.026 & .176\,\textpm\,.027 & .200\,\textpm\,.030 & 42.02\,\textpm\,7.59 & .016\,\textpm\,.002 \\
\bottomrule
\end{tabular}}%
\caption{Text-encoder ablation for Ours-Lite with the Split-branch RVQ +
SemDist tokenizer. Panel (a) uses the HumanML3D HML-263 evaluator with batch
size 32; panel (b) uses TMR-SOMA on the overall test set with 256 clips per
repeat. Values are mean $\pm$ standard deviation over 30 repeats.}
\label{tab:ab_t2m_text}
\end{table*}

SigLIP produces the lowest HML-263 FID, whereas Flan-T5 is stronger on the
overall TMR-SOMA retrieval track and is used for the main models. Qwen3 is
mid-ranked on HML-263 but last on TMR-SOMA, where its median rank is 42.02
against 1.70 for Flan-T5. The two tracks therefore order the three encoders
differently.

\subsection{Source-wise Tokenizer Ablation}
\label{sup:tokenizer-per-source}

The main paper compares tokenizer variants using the full-test Overall pool.
Table~\ref{tab:ab_t2m_tokenizer_per_dataset} resolves that comparison over the
three provenance subsets shared by all four variants.
\begin{table*}[t]
\centering
\small
\setlength{\tabcolsep}{2.3pt}
\adjustbox{max width=\linewidth}{%
\begin{tabular}{@{}lllrrrrrr@{}}
\toprule
Subset & RVQ layout & Sem. & FID$\downarrow$ & R@1$\uparrow$ & R@2$\uparrow$ &
R@3$\uparrow$ & R@5$\uparrow$ & MedR$\downarrow$ \\
\midrule
\multirow{4}{*}{HumanML3D}
& Single-chain & $\times$ & \textbf{.919\,\textpm\,.094}
  & \underline{.381\,\textpm\,.110} & \underline{.522\,\textpm\,.126}
  & \underline{.637\,\textpm\,.109} & \underline{.779\,\textpm\,.123}
  & \underline{2.40\,\textpm\,1.00} \\
& Single-chain & $\checkmark$ & 1.006\,\textpm\,.092
  & \textbf{.395\,\textpm\,.097} & \textbf{.583\,\textpm\,.103}
  & \textbf{.686\,\textpm\,.077} & \textbf{.787\,\textpm\,.077}
  & \textbf{2.07\,\textpm\,.61} \\
& Split-branch & $\times$ & .977\,\textpm\,.092
  & .227\,\textpm\,.085 & .365\,\textpm\,.101
  & .462\,\textpm\,.126 & .605\,\textpm\,.099
  & 4.28\,\textpm\,1.28 \\
& Split-branch & $\checkmark$ & \underline{.920\,\textpm\,.101}
  & .326\,\textpm\,.103 & .451\,\textpm\,.084
  & .540\,\textpm\,.107 & .707\,\textpm\,.115
  & 3.23\,\textpm\,1.16 \\
\midrule
\multirow{4}{*}{BONES-SEED}
& Single-chain & $\times$ & \underline{.237\,\textpm\,.018}
  & \underline{.512\,\textpm\,.043} & \underline{.646\,\textpm\,.043}
  & \underline{.702\,\textpm\,.039} & \underline{.757\,\textpm\,.040}
  & \textbf{1.35\,\textpm\,.48} \\
& Single-chain & $\checkmark$ & .249\,\textpm\,.017
  & .489\,\textpm\,.037 & .615\,\textpm\,.030
  & .672\,\textpm\,.031 & .734\,\textpm\,.039
  & 1.55\,\textpm\,.50 \\
& Split-branch & $\times$ & .269\,\textpm\,.018
  & .366\,\textpm\,.040 & .481\,\textpm\,.042
  & .549\,\textpm\,.042 & .617\,\textpm\,.044
  & 2.83\,\textpm\,.87 \\
& Split-branch & $\checkmark$ & \textbf{.219\,\textpm\,.016}
  & \textbf{.518\,\textpm\,.039} & \textbf{.654\,\textpm\,.035}
  & \textbf{.715\,\textpm\,.045} & \textbf{.775\,\textpm\,.043}
  & \textbf{1.35\,\textpm\,.48} \\
\midrule
\multirow{4}{*}{MotionGV}
& Single-chain & $\times$ & \textbf{.795\,\textpm\,.100}
  & \underline{.712\,\textpm\,.096} & .851\,\textpm\,.054
  & .905\,\textpm\,.053 & .948\,\textpm\,.050
  & \textbf{1.00\,\textpm\,.00} \\
& Single-chain & $\checkmark$ & \underline{.796\,\textpm\,.102}
  & \textbf{.721\,\textpm\,.101} & \textbf{.866\,\textpm\,.073}
  & \underline{.916\,\textpm\,.068} & \textbf{.958\,\textpm\,.040}
  & \underline{1.03\,\textpm\,.18} \\
& Split-branch & $\times$ & .823\,\textpm\,.089
  & .602\,\textpm\,.117 & .762\,\textpm\,.114
  & .850\,\textpm\,.075 & .921\,\textpm\,.065
  & 1.18\,\textpm\,.38 \\
& Split-branch & $\checkmark$ & .805\,\textpm\,.094
  & \underline{.712\,\textpm\,.104} & \underline{.854\,\textpm\,.085}
  & \textbf{.924\,\textpm\,.054} & \underline{.949\,\textpm\,.038}
  & \textbf{1.00\,\textpm\,.00} \\
\bottomrule
\end{tabular}}%
\caption{Source-wise TMR-SOMA tokenizer ablation with matched Ours-Lite Flan-T5
packet generators. Values are mean $\pm$ sample standard deviation over 30
repeats. We report the three provenance subsets with retained outputs for all
four variants; HumanSC3D is omitted because its Plain RVQ per-subset output is
unavailable.}
\label{tab:ab_t2m_tokenizer_per_dataset}
\end{table*}

The aggregate advantage of the split semantic route is most consistent on
BONES-SEED, where it obtains the best FID and every retrieval rate and ties the
best median rank. The HumanML3D subset instead
favors Plain RVQ in FID and semantic Single-chain RVQ in retrieval, while the
MotionGV ranking is mixed across metrics.

\subsection{Full Tokenizer Reconstruction Ablation}
\label{sup:tokenizer-recon}

Table~\ref{tab:ab_recon_tokenizer} reports the complete matched reconstruction
metrics for the four tokenizer variants and a source-wise breakdown for the
retained decomposed runs.
\begin{table*}[t]
\centering
\setlength{\tabcolsep}{4pt}
\adjustbox{max width=\linewidth}{%
\begin{tabular}{@{}llrrrr@{}}
\toprule
\multicolumn{6}{c}{(a) Full test split} \\
\midrule
RVQ layout & SemDist. & MPJPE-22$\downarrow$ & Median$\downarrow$ & MPJPE-77$\downarrow$ & PA-MPJPE$\downarrow$ \\
\midrule
Single-chain & $\times$ & \underline{11.98} & \underline{9.28} & \underline{15.60} & \underline{10.12} \\
Single-chain & $\checkmark$ & 13.18 & 10.44 & 18.54 & 10.84 \\
Split-branch & $\times$ & \textbf{10.92} & \textbf{9.08} & \textbf{13.70} & \textbf{9.58} \\
\textbf{Split-branch} & $\checkmark$ & 12.83 & 10.13 & 15.93 & 11.07 \\
\bottomrule
\end{tabular}}%
\par
\adjustbox{max width=\linewidth}{%
\begin{tabular}{@{}lrlcrrr@{}}
\toprule
\multicolumn{7}{c}{(b) Source-wise reconstruction} \\
\midrule
Source & $n$ & RVQ layout & SemDist. & MPJPE-22$\downarrow$ & MPJPE-77$\downarrow$ & PA-MPJPE$\downarrow$ \\
\midrule
\multirow{2}{*}{BONES-SEED} & \multirow{2}{*}{52,713} & Single-chain & $\times$ & \textbf{9.32} & 12.61 & \textbf{8.03} \\
& & \textbf{Split-branch} & $\checkmark$ & 9.61 & \textbf{12.60} & 8.40 \\
\multirow{2}{*}{MotionGV} & \multirow{2}{*}{81,418} & Single-chain & $\times$ & \textbf{13.52} & \textbf{17.21} & \textbf{11.31} \\
& & \textbf{Split-branch} & $\checkmark$ & 14.73 & 17.78 & 12.65 \\
\multirow{2}{*}{HumanML3D} & \multirow{2}{*}{2,108} & Single-chain & $\times$ & \textbf{18.69} & \textbf{27.23} & \textbf{15.94} \\
& & \textbf{Split-branch} & $\checkmark$ & 19.22 & 27.57 & 16.42 \\
\bottomrule
\end{tabular}}%
\caption{Motion-tokenizer reconstruction ablation on the v260717 test split.
Panel (a) reports all four variants on 136,480 clips; panel (b) reports
source-wise results for the two variants with retained decomposed evaluations.
All variants use the same 50 Hz motion representation; errors are in mm.}
\label{tab:ab_recon_tokenizer}
\end{table*}

Semantic supervision raises MPJPE-77 by 2.23\,mm in the split layout, from
13.70 to 15.93\,mm, and by 2.94\,mm in the single chain, from 15.60 to
18.54\,mm. The split layout is also the more accurate of the two under both
settings. Panel (b) resolves the two retained variants by provenance; their
MPJPE-77 differs by 0.01\,mm on BONES-SEED, 0.57\,mm on MotionGV, and
0.34\,mm on HumanML3D.

\subsection{Semantic Branch Design}
\label{sup:sem-branch}

Table~\ref{tab:ab_sem_branch} studies where the semantic constraint is applied
at a fixed weight, reporting reconstruction together with the agreement between
the semantic representation and the frozen teacher. An unsupervised run gives
the reconstruction reference.
\begin{table*}[!htbp]
\centering
\setlength{\tabcolsep}{4pt}
\adjustbox{max width=\linewidth}{%
\begin{tabular}{@{}lrrrr@{}}
\toprule
Semantic route & MPJPE-77$\downarrow$ & Median$\downarrow$
& Cos$\uparrow$ & R@10$\uparrow$ \\
\midrule
No semantic distillation & \textbf{12.78} & \textbf{10.83} & -- & -- \\
\midrule
\textbf{Split-branch} & \underline{13.18} & \underline{11.27} & \textbf{.878} & \textbf{.672} \\
Single-chain, layer 0 & 15.43 & 13.37 & .770 & .543 \\
Single-chain, layers 0--1 & 15.85 & 13.68 & \underline{.846} & \underline{.627} \\
\bottomrule
\end{tabular}}%
\caption{Placement of the semantic constraint, with
$\lambda_{\mathrm{sem}}=0.15$ in every supervised run. All runs are trained on
BONES-SEED (142{,}000 clips) for 1{,}000 epochs and share the encoder, decoder,
codebook sizes, and optimization schedule. Reconstruction errors are in mm over
its own 14{,}198-clip test split. Cos is the cosine similarity between the
aggregated semantic representation and the frozen teacher embedding. Retrieval
uses an 11{,}547-clip pool, for which chance R@10 is $8.7\times10^{-4}$.}
\label{tab:ab_sem_branch}
\end{table*}

Routing the constraint to a parallel branch costs 0.40\,mm against that
reference, whereas applying it to the first residual level of a single chain
costs 2.65\,mm and to the first two levels 3.07\,mm. The parallel branch also
reaches the closest agreement with the teacher, at .878 cosine similarity and
.672 R@10, against .770 and .543 for the single level and .846 and .627 for the
first two levels. Placing the constraint outside the residual chain therefore
buys semantic alignment at a smaller reconstruction cost than placing it
inside.

\begin{table*}[!t]
\centering
\small
\setlength{\tabcolsep}{1mm}
\adjustbox{max width=\linewidth}{%
\begin{tabular}{@{}lrrrrll@{}}
\toprule
Source & Pairs & Groups & Hours & Mean (s) & Motion & Text \\
\midrule
MotionGV & 542,787 & 542,787 & 690.7 & 4.58 & Monocular video & GPT-4o \\
BONES-SEED & 351,422 & 141,869 & 281.2 & 2.88 & Optical marker & Dataset timeline \\
HumanML3D & 14,094 & 12,886 & 28.1 & 7.18 & Optical marker & Manual \\
Fit3D & 922 & 922 & 5.1 & 19.92 & Optical marker & MotionHub \\
HumanSC3D & 688 & 688 & 0.9 & 4.87 & Optical marker & MotionHub \\
\midrule
Total & 909,913 & 699,152 & 1,006.0 & 3.98 & & \\
\bottomrule
\end{tabular}}%
\caption{Composition of $\Omega$-MotionVerse. Pairs counts text--motion pairs;
Groups counts the source recordings used as the atomic unit of the split, so
pairs cut from one recording never separate. Every subset is partitioned
80:5:15 over these groups, except HumanML3D, which keeps its official split;
the Kimodo benchmark test portion of BONES-SEED is forced into the test set.}
\label{tab:motionverse_composition}
\end{table*}

\section{Implementation and Data Details}
\label{sup:implementation}

\subsection{$\Omega$-MotionVerse Composition and Splits}
\label{sup:motionverse-composition}

Table~\ref{tab:motionverse_composition} reports the per-source composition of
the corpus. MotionGV and BONES-SEED together account for 98.3\% of all
text--motion pairs. The corpus contains 909,913 pairs over 699,152 source
groups, approximately 1,006 hours of motion, and an average clip duration of
3.98 seconds.

\paragraph{Source selection and conversion.}
MotionGV is taken from the MotionMillion distribution~\cite{fan2025go}; the
other MotionMillion subsets are excluded when they have unreliable
text--motion alignment, game-asset provenance, severe reconstruction noise, or
missing captions. It contains motions reconstructed from in-the-wild monocular
video, which we convert from their native 272-D representation through SMPL-X
and SOMA-X into the common SOMA skeleton. When several captions are available,
the most complete caption is retained. Fit3D~\cite{fieraru2021aifit} and
HumanSC3D~\cite{fieraru2021learning} contain studio marker captures with
MotionHub captions~\cite{ling2025versatilemotionunifiedframeworkmotion} and
follow the same conversion path.

BONES-SEED~\cite{luo2026sonicsupersizingmotiontracking} already provides
optical Vicon recordings on the SOMA skeleton. Its long recordings are
segmented into action clips using the dataset timeline annotations. HumanML3D
motions~\cite{guo2022generating} are recovered from AMASS and HumanAct12
through SMPL-X and SOMA-X pose inversion. Official frame ranges are used,
mirrored examples are excluded, and multiple valid captions for one retained
motion remain distinct text--motion pairs.

\paragraph{Common motion standardization.}
Every source is converted to the SOMA77 skeleton convention, resampled to
50 Hz, floor aligned, and canonicalized at its first frame. Canonicalization
removes the initial planar translation and heading while retaining a separate
anchor containing the initial absolute pose and root transform. The
standardized UMR499 frame representation contains root-trajectory features,
root and non-root 6-D rotations, sparse joint velocities, and foot-contact
states. The token stream models motion evolution, while the anchor supplies
the boundary condition used during decoding and forward kinematics.

\paragraph{Split policy.}
Splits are drawn over underlying source recordings rather than individual
clips, so all clips cut from one recording remain in the same partition. The
split seed is 20260717. The resulting corpus contains 727,941 training pairs,
45,492 validation pairs, and 136,480 test pairs, holding 804.6, 50.4, and
151.1 hours of motion; Table~\ref{tab:motionverse_splits} resolves these
counts per source. Most sources use an 80:5:15 group-level split, while
HumanML3D retains its official split. Its test subset contains 2,108 pairs
over 1,921 unique clips and covers 87.6\% of the 2,192 non-mirrored official
test clips.
\begin{table}[t]
\centering
\small
\setlength{\tabcolsep}{1mm}
\adjustbox{max width=\linewidth}{%
\begin{tabular}{@{}lrrrr@{}}
\toprule
Source & Train & Val & Test & Total \\
\midrule
MotionGV & 434,230 & 27,139 & 81,418 & 542,787 \\
BONES-SEED & 281,135 & 17,574 & 52,713 & 351,422 \\
HumanML3D & 11,287 & 699 & 2,108 & 14,094 \\
Fit3D & 738 & 46 & 138 & 922 \\
HumanSC3D & 551 & 34 & 103 & 688 \\
\midrule
Total & 727,941 & 45,492 & 136,480 & 909,913 \\
\bottomrule
\end{tabular}}%
\caption{Text--motion pairs per source and partition.}
\label{tab:motionverse_splits}
\end{table}

For BONES-SEED, the complete Kimodo benchmark test portion is forced into the
test split. It contains 13,884 recordings and 33,970 pairs, accounting for
64.4\% of the BONES-SEED test pairs. No Kimodo test recording appears in the
training or validation partitions.

\paragraph{Deduplication.}
Each sample receives a deterministic identifier, and repeated submissions of
the same motion are removed using content hashes. HumanML3D is the deliberate
pair-level exception: its 14,094 text--motion pairs correspond to 12,886 unique
clips because one motion may retain several captions. Clip-level statistics
therefore deduplicate HumanML3D rows by clip identifier. Caption multiplicity
for one retained motion is not treated as motion duplication.

\subsection{TMR-SOMA Training and Use}
\label{sup:tmr-soma}

TMR-SOMA is a motion--text retrieval model and evaluation space rather than a
text-to-motion generator. It has two separate roles: a frozen motion encoder
provides semantic teacher descriptors for SeMoCo, and a training-split-adapted
then frozen copy defines the TMR-SOMA evaluation space.

\paragraph{Inputs and architecture.}
Adaptation uses paired samples from the $\Omega$-MotionVerse training
partition, with validation pairs for checkpoint selection. SOMA77 joints are
mapped to the evaluator's
30-joint SOMA skeleton, canonicalized, normalized using the pretrained motion
statistics, and cached as 186-D frame features in fp16. Text inputs are cached
2,048-D token-level Flan-T5-XL features. Text and motion sequences are padded
independently with validity masks, and examples with fewer than two motion
packets are excluded from paired adaptation.

The evaluator follows the TMR retrieval formulation~\cite{petrovich2023tmr}
and uses paired ACTOR-style variational Transformer encoders for motion and
text. Each encoder has latent width 256, six Transformer layers, four attention
heads, feed-forward width 1,024, dropout 0.1, and GeLU activation. Each encoder
produces the mean and log variance of a 256-D diagonal Gaussian; a latent sample
is drawn by reparameterization and L2-normalized for retrieval. A six-layer
Transformer motion decoder maps a latent back to the 186-D motion-feature
sequence and acts as a geometric anchor for the shared space.

\paragraph{Initialization and adaptation.}
The motion encoder and decoder are initialized from TMR-SOMA-RP-v1, the model
released with Kimodo~\cite{rempe2026kimodo}. The text encoder uses the same
ACTOR-style architecture but accepts 2,048-D Flan-T5-XL token features.
Compatible Transformer weights are copied from the pretrained motion encoder,
while the modality-specific text projection is initialized for the new input
width. During joint adaptation, both encoders are trainable and the pretrained
motion decoder is frozen.

Let the text and motion encoders produce Gaussian parameters
$(\boldsymbol{\mu}_t,\log\boldsymbol{\sigma}_t^2)$ and
$(\boldsymbol{\mu}_m,\log\boldsymbol{\sigma}_m^2)$. Reparameterized samples are
L2-normalized and aligned with a symmetric InfoNCE objective using a learnable
logit scale initialized at temperature 0.1. The full objective combines motion
reconstruction with weight 1.0, symmetric contrastive alignment with weight
0.1, SmoothL1 alignment between latent means with weight $10^{-5}$, KL
regularization toward a unit Gaussian with weight $10^{-5}$, and bidirectional
cross-modal KL matching with weight $10^{-5}$. Latents from both modalities
are decoded through the fixed motion decoder.

\paragraph{Optimization and freezing.}
TMR-SOMA adaptation uses AdamW with learning rate $10^{-4}$, zero weight decay,
1,000 linear warmup steps followed by cosine decay, gradient clipping at 1.0,
bf16 precision, and random seed 3407. The per-process batch size is 256,
training runs for 200,000 optimization steps, validation occurs every 1,000
steps, and checkpoints are saved every 5,000 steps. The selected text and
motion encoders are then exported and frozen.

\subsection{SeMoCo Architecture and Training Details}
\label{sup:semoco-training}

SeMoCo receives 50-Hz UMR499 transition records and uses a stride-four
convolutional encoder and decoder with dilated residual blocks. It emits one
motion packet at 12.5 Hz. Each packet is
\begin{equation}
  \mathbf{m}_t =
  [q_t^{\mathrm{sem}}, q_t^{\mathrm{kin},1}, \ldots,
   q_t^{\mathrm{kin},15}],
  \label{eq:sup-motion-packet}
\end{equation}
where the first code is produced by a semantic vector-quantization branch and
the other 15 codes by a kinematic residual vector quantizer. Both branches
receive the full encoder latent through separate learned projections. Their
quantized outputs are projected into a shared decoder space and summed before
motion reconstruction.
The encoder has latent width 512 and three residual blocks per stage. The
semantic codebook and all 15 kinematic codebooks contain 1,024 entries. Their
EMA coefficient is 0.99, and quantizer dropout with probability 0.2 varies the
number of active kinematic levels while always retaining the semantic code,
which teaches the decoder to reconstruct motion from progressively refined
packets.

\paragraph{Sampling and semantic supervision.}
Training uses 64-frame windows, corresponding to 1.28 seconds or 16 motion
packets. Windows are sampled with stride 32, and at most four windows are
selected from one source clip in a training sample. The semantic target is a
256-D unit-normalized descriptor from the frozen TMR-SOMA-RP-v1 motion encoder.
One stop-gradient target is computed for each window, and a temporal semantic
head aggregates its sequence of 16 quantized semantic embeddings. This cosine
alignment organizes the first code around behavior-level information.

\paragraph{Reconstruction objective.}
The decoder is supervised in motion and geometry space using joint-position
reconstruction, first-order velocity consistency, second-order acceleration
consistency, horizontal foot velocity at ground-truth contact frames, and the
usual codebook and commitment losses. The auxiliary weights are
\begin{equation}
\begin{split}
  \lambda_{\mathrm{vel}}&=0.5, \qquad
  \lambda_{\mathrm{acc}}=0.25, \qquad
  \lambda_{\mathrm{skate}}=0.5, \\
  \lambda_{\mathrm{VQ}}&=0.02, \qquad
  \lambda_{\mathrm{sem}}=0.15.
\end{split}
\label{eq:sup-semoco-weights}
\end{equation}

SeMoCo is optimized with AdamW using weight decay $10^{-4}$, betas
$(0.9,0.95)$, 500 linear warmup steps followed by cosine decay, gradient
clipping at 1.0, bf16 precision, and seed 3407. It is trained with a batch of
1,024 per process on eight processes, and the reported generators use its
best-validation checkpoint. The selected
tokenizer is then frozen and every corpus clip is encoded once into a cached
packet sequence. Each tokenizer ablation uses the packets and reconstructions
produced by that same tokenizer.

\subsection{Dual-Axis Motion Transformer Learning}
\label{sup:generator-learning}

\begin{table*}[t]
\centering
\small
\setlength{\tabcolsep}{4.5pt}
\begin{tabular}{lccc}
\toprule
Component & Ours-Lite & Ours-Base & Interpretation \\
\midrule
Trainable parameters & 188.0M & 391.1M & Frozen text encoder excluded \\
Temporal layers & 12 & 24 & Base doubles temporal depth \\
Temporal hidden width & 768 & 1,024 & Base uses a wider temporal state \\
Temporal query/KV heads & 6 / 3 & 8 / 4 & 128-D heads and 2:1 GQA \\
Temporal FFN width & 2,048 & 2,730 & Scaled with backbone width \\
Code-predictor width/layers & 1,024 / 5 & 1,024 / 5 & Fixed across sizes \\
Code-predictor query/KV heads & 8 / 4 & 8 / 4 & Fixed across sizes \\
Code-predictor FFN width & 3,072 & 3,072 & Fixed across sizes \\
Backbone-to-code bridge & Learned $768\!\rightarrow\!1{,}024$ & Identity & Width conversion only for Lite \\
Packet structure/rate & \multicolumn{2}{c}{16 codes at 12.5 Hz} & Fixed representation \\
Text encoder & \multicolumn{2}{c}{Frozen Flan-T5-XL} & Fixed conditioning source \\
\bottomrule
\end{tabular}
\caption{Ours-Lite and Ours-Base instantiate the same dual-axis generator at
different temporal capacities. Parameter counts cover the trainable generator
and exclude the frozen text encoder. Both variants are trained for 100{,}000
updates; reported results use the best validation checkpoint among the
1{,}000-step validation saves, which for both variants is step 99{,}000.}
\label{tab:sup-model-variants}
\end{table*}

Ours-Base and Ours-Lite are two capacity variants of the same generator. They
consume the same 16-code packets at 12.5 Hz, use the same codebook vocabularies
and ordering, and reconstruct motion through the same frozen SeMoCo decoder.
Scaling therefore changes temporal model capacity rather than motion
resolution or tokenizer reconstruction fidelity.

\paragraph{Temporal-axis modeling.}
Each codebook has a separate embedding table. The 16 embeddings at one time
step are summed to form a single packet embedding, so the temporal sequence
remains at 12.5 Hz rather than expanding by a factor of 16. The temporal
Transformer processes the text condition, a learned motion-BOS token, and the
preceding completed packets. Its next-interval hidden state directly predicts
the semantic code $q_t^{\mathrm{sem}}$ and a binary EOS target. This axis models
long-range action progression, duration, ordering, and consistency between the
description and motion history.

\paragraph{Codebook-axis modeling.}
After the semantic code is selected, a lightweight code predictor (causal
depth decoder) generates the 15 kinematic residual codes autoregressively:
\begin{equation}
  q_t^{\mathrm{sem}} \rightarrow q_t^{\mathrm{kin},1}
  \rightarrow \cdots \rightarrow q_t^{\mathrm{kin},15}.
\end{equation}
At every step it is conditioned on the temporal hidden state and the preceding
codes of the current packet. The code predictor is fixed across model sizes:
it has width 1,024, five layers, eight query heads, four KV heads, and
feed-forward width 3,072. Ours-Lite uses a learned $768\!\rightarrow\!1{,}024$
bridge into this module, whereas Ours-Base uses an identity connection.

\paragraph{Text conditioning and shared conventions.}
Both variants use cached token-level features from a frozen Flan-T5-XL text
encoder. The 2,048-D features are normalized and projected to the temporal
backbone width. Text tokens attend bidirectionally within the text prefix;
motion tokens attend to all valid text tokens and causally to preceding motion
tokens; text queries do not attend to motion tokens. A learned null-text token
supports classifier-free guidance, and a learned motion-BOS token begins the
packet sequence.

Both variants use RMSNorm, SwiGLU feed-forward layers, rotary position
embeddings, QK normalization, grouped-query attention with a 2:1 query-to-KV
head ratio, attention-head dimension 128, RoPE base 1,000,000, maximum sequence
length 4,096, and zero generator dropout.

\paragraph{Training data and context budget.}
Generator training begins after SeMoCo is frozen. Each example contains a
cached 16-code packet sequence, cached Flan-T5-XL token features, valid text and
motion lengths, and an EOS target. Text is capped at 64 tokens and motion at
300 packets, or approximately 24 seconds at 12.5 Hz, within a 372-position
training context. Each example retains its actual valid text and motion
lengths.

\paragraph{Teacher forcing and objective.}
Teacher forcing is applied along both axes. The temporal Transformer receives
ground-truth completed packets as history and predicts the next semantic code;
the code predictor receives the ground-truth prefix of the target packet and
predicts its next residual code. The objective is
\begin{equation}
\begin{split}
  \mathcal{L}_{\mathrm{gen}} ={}&
  \lambda_0 \sum_t \mathcal{L}_{\mathrm{CE}}^{\mathrm{sem}}(t)
  + \lambda_{\mathrm{code}}
  \sum_t \sum_{i=1}^{15}
  \lambda_i \mathcal{L}_{\mathrm{CE}}^{\mathrm{kin},i}(t) \\
  &+ \lambda_{\mathrm{eos}}
  \sum_t \mathcal{L}_{\mathrm{EOS}}(t).
\end{split}
\label{eq:sup-generator-objective}
\end{equation}
The semantic code has weight 1.5; residual level 1 has weight 1.2, levels 2--11
have weight 1.0, and levels 12--15 have weight 0.7. The EOS loss has weight
1.0. The combined code-axis term uses $\lambda_{\mathrm{code}}=0.3$ and is
linearly warmed up over the first 5,000 steps, allowing temporal semantic
prediction to stabilize before the full within-packet objective reaches its
target scale.

Text conditioning is replaced by the learned null-text token with probability
0.1 during training. This jointly learns conditional and unconditional
predictions for classifier-free guidance at inference time.

\begin{table*}[t]
\centering
\small
\setlength{\tabcolsep}{5pt}
\begin{tabular}{ll@{\qquad}ll}
\toprule
Setting & Value & Setting & Value \\
\midrule
Optimizer & AdamW & Learning rate & $2\times10^{-4}$ \\
Weight decay & 0.01 & Adam betas & $(0.9, 0.95)$ \\
LR schedule & 4,000-step warmup + cosine & Gradient clipping & 1.0 \\
Precision & bf16 & Batch size & 32 per process \\
Gradient accumulation & None & Optimization steps & 100,000 \\
Data-parallel processes & 1 (Lite) / 2 (Base) & Global batch size & 32 (Lite) / 64 (Base) \\
Validation interval & 1,000 steps & Checkpoint interval & 5,000 steps \\
Random seed & 3407 & Text-condition dropout & 0.1 \\
Maximum text length & 64 tokens & Maximum motion length & 300 packets \\
Training context & 372 positions & Residual-loss warmup & 5,000 steps \\
\bottomrule
\end{tabular}
\caption{Generator training configuration. Ours-Lite and Ours-Base share the
same data, objective, optimizer settings, update count, tokenizer, code
predictor, and text encoder, and use the same per-process batch of 32.
Ours-Lite is trained on one data-parallel process and Ours-Base on two, giving
global batch sizes of 32 and 64.}
\label{tab:sup-generator-training}
\end{table*}

The complete training order is: (1) standardize motions and construct splits;
(2) adapt and freeze TMR-SOMA using only the training partition; (3) train
SeMoCo with frozen semantic teacher descriptors; (4) freeze SeMoCo and cache
packets; (5) train Ours-Lite and Ours-Base independently on the same cached
pairs; and (6) freeze all models and evaluate held-out generations in the
appropriate fixed evaluator space. No gradient crosses a stage boundary.

\subsection{Motion Prediction}
\label{sup:motion-prediction}

Motion-prediction models are trained independently from the text-conditioned
generators. The observed prefix is encoded by SeMoCo into completed packets
and inserted directly into causal temporal history. The predictor uses the
same semantic-first temporal and codebook factorization but receives neither
text features nor a learned null-text surrogate. During training, the observed
prefix remains fixed while future packet histories and within-packet prefixes
are supplied through teacher forcing. We train two capacity variants on this
task, Ours-Lite with approximately 150M parameters and Ours-Base with
approximately 400M parameters. Both use the same tokenizer, objective,
optimizer settings, 100{,}000-step training budget, and 2{,}048-packet
context; only temporal backbone capacity differs. Each variant is trained
with AdamW, learning rate $2.5\times10^{-4}$, weight decay 0.1, 1{,}000 linear
warmup steps followed by cosine decay, gradient clipping at 1.0, bf16
precision, and seed 3407.

\subsection{Computing Infrastructure}
\label{sup:infrastructure}

Experiments are run on single-node servers, each equipped with eight NVIDIA
H100 80\,GB GPUs, 96 CPU cores, and 1\,TB of system memory. The software
environment is Ubuntu 22.04 LTS with CUDA 12.6, and training runs in bf16
precision through \texttt{torch.distributed}. SeMoCo is trained under Python
3.12 with PyTorch 2.12.0, and the generators under Python 3.11 with PyTorch
2.13.0. Flan-T5-XL text features are precomputed offline, so no language model
is loaded during generator training.

\subsection{Evaluation Protocols}
\label{sup:evaluation}

\paragraph{Evaluator spaces.}
HumanML3D results use its official evaluator~\cite{guo2022generating} and
batch-32 retrieval; we refer to this feature space as HML-263. TMR-SOMA is
initialized from the evaluator released with Kimodo~\cite{rempe2026kimodo},
adapted only on the $\Omega$-MotionVerse training partition, and frozen before
evaluation. The two spaces use different motion representations, embedding
models, and normalizations. Native SOMA outputs are evaluated directly; SMPL
or HumanML3D outputs are first retargeted to SOMA and marked accordingly.

\paragraph{TMR-SOMA repeats.}
Each provenance result uses 256 clips per repeat and reports mean and standard
deviation over 30 repeats.

\paragraph{FootSkate.}
FootSkate is the mean horizontal foot-joint speed in meters per second over
heuristically detected contact frames. We use the left and right ankle and toe
joints of the SOMA77 skeleton;
a foot is considered in contact when its 3-D speed is below 0.15 m/s and its
height is below 0.12 m. The reported value averages the horizontal $x$--$z$
speed over all detected contact frames, and lower values indicate less foot
sliding.

\paragraph{Jerk.}
Let $\mathbf{p}_{t,j}\in\mathbb{R}^{3}$ denote the position of joint $j$ at
frame $t$, and let $\Delta t=1/f$ for evaluation frame rate $f$. We approximate
the third temporal derivative as
\begin{equation}
\mathbf{J}_{t,j}
=
\frac{
\mathbf{p}_{t+3,j}-3\mathbf{p}_{t+2,j}
+3\mathbf{p}_{t+1,j}-\mathbf{p}_{t,j}
}{(\Delta t)^3}.
\label{eq:sup-jerk}
\end{equation}
The reported motion smoothness is the mean joint-jerk magnitude,
\begin{equation}
\operatorname{Jerk}
=
\frac{1}{(T-3)J}
\sum_{t=0}^{T-4}\sum_{j=1}^{J}
\left\|\mathbf{J}_{t,j}\right\|_2,
\label{eq:sup-mean-jerk}
\end{equation}
where $T$ is the number of frames and $J$ is the number of evaluated joints.
We compute the metric on decoded joint positions after the coordinate conversion
and resampling of the corresponding protocol, without additional smoothing.
Its unit is $\mathrm{m/s^3}$, and lower values indicate smoother motion.

\paragraph{Reconstruction.}
Decoded output is converted to joints and pelvis aligned before MPJPE;
PA-MPJPE adds a per-sequence similarity alignment. MPJPE-22 and MPJPE-77 denote
the metric over the 22 SMPL body joints and over all 77 SOMA joints.

\paragraph{Motion prediction.}
Motion prediction observes the first 0.5 seconds and evaluates the following
two seconds in each method's native pipeline, using native valid pools and
frame rates. ADE and FDE are measured over the predicted segment, while
best-of-50 reports the lowest error among 50 independent samples for each clip.

\end{document}